\documentclass[acmtog,authorversion]{acmart}
\usepackage{xcolor}
\renewcommand\footnotetextcopyrightpermission[1]{} 
\usepackage{subcaption}
\usepackage{xspace}
\makeatletter
\DeclareRobustCommand\onedot{\futurelet\@let@token\@onedot}
\def\@onedot{\ifx\@let@token.\else.\null\fi\xspace}

\def\eg{\emph{e.g}\onedot} 
\def\ie{\emph{i.e}\onedot}

\makeatletter
\def\runningfoot{\def\@runningfoot{}}
\def\firstfoot{\def\@firstfoot{}}
\makeatother
\usepackage[ruled]{algorithm2e} 

\SetAlFnt{\small}
\SetAlCapFnt{\small}
\SetAlCapNameFnt{\small}
\SetAlCapHSkip{0pt}

\acmJournal{TOG}

\begin{document}

\title{
Inverse Painting:  Reconstructing The Painting Process
}

\author{Bowei Chen}
\orcid{0000-0002-2225-8796}
\affiliation{%
 \institution{University of Washington}
 \streetaddress{1410 NE Campus Pkwy}
 \city{Seattle}
 \state{WA}
 \postcode{98195}
 \country{USA}}
\email{boweiche@cs.washington.edu}
\author{Yifan Wang}
\orcid{0000-0002-8246-2254}
\affiliation{%
 \institution{University of Washington}
 \city{Seattle}
 \country{USA}
}
\email{yifan1@cs.washington.edu}
\author{Brian Curless}
\orcid{0000-0002-0095-5400}
\affiliation{%
 \institution{University of Washington}
 \city{Seattle}
 \country{USA}
}
\email{curless@cs.washington.edu}
\author{Ira Kemelmacher-Shlizerman}
\orcid{0009-0003-9498-584X}
\affiliation{%
 \institution{University of Washington}
 \city{Seattle}
 \country{USA}
}
\email{kemelmi@cs.washington.edu}
\author{Steven M. Seitz}
\orcid{0009-0000-4214-4078}
\affiliation{%
 \institution{University of Washington}
 \city{Seattle}
 \country{USA}
}
\email{seitz@cs.washington.edu}

\renewcommand\shortauthors{Chen, et al}

\begin{abstract}
Given an input painting, we reconstruct a time-lapse video of how it may have been painted.  We formulate this as an autoregressive image generation problem, in which an initially blank ``canvas'' is iteratively updated.  The model learns from real artists by training on many painting videos.
Our approach incorporates text and region understanding to define a set of painting ``instructions'' and updates the canvas with a novel diffusion-based renderer. The method extrapolates beyond the limited, acrylic style paintings on which it has been trained, showing plausible results for a wide range of artistic styles and genres.
Our project page and code are available at:
\color{magenta}{\url{https://inversepainting.github.io/}}
\end{abstract}

\begin{teaserfigure}
        \centering
    \captionsetup{type=figure}
    \includegraphics[scale=0.73]{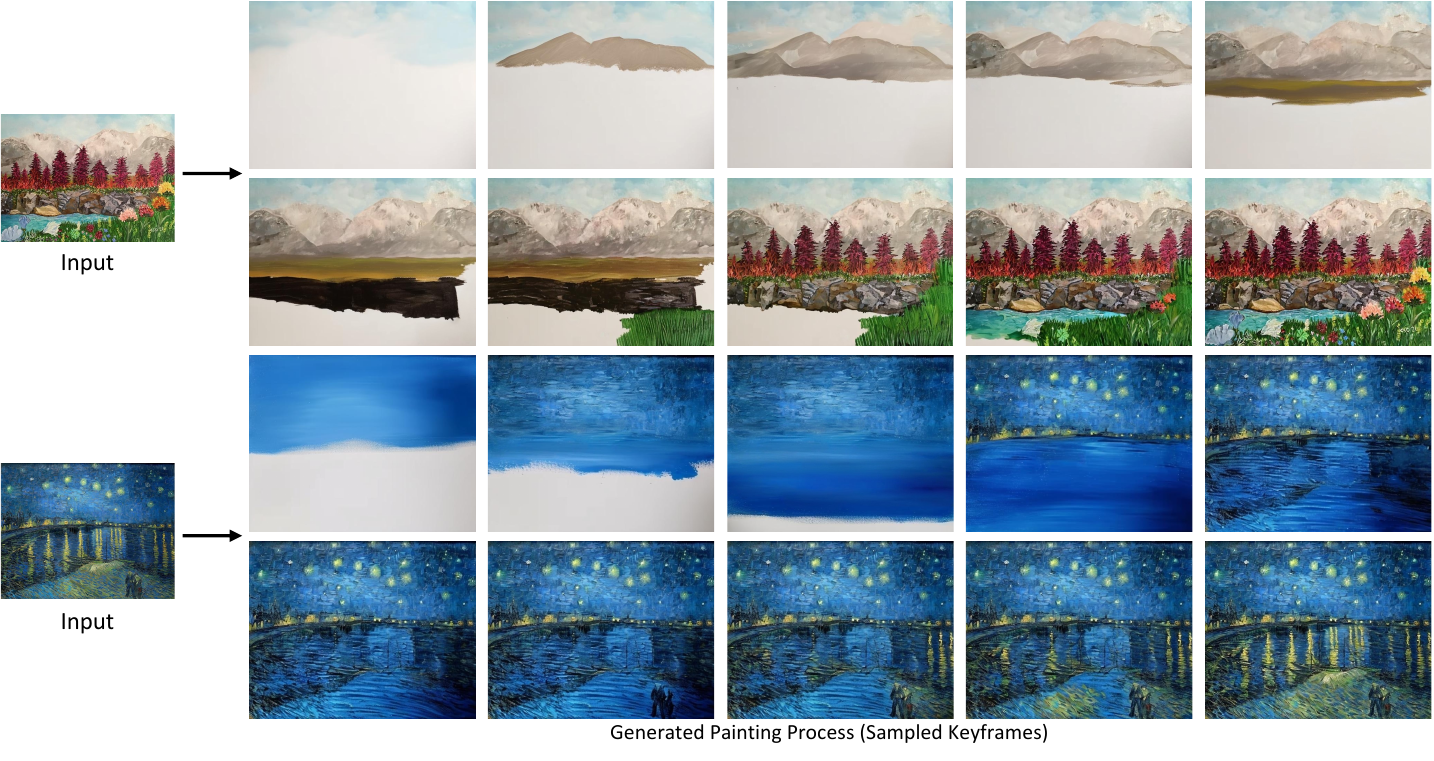}
        \captionof{figure}{We present Inverse Painting, a diffusion-based method to generate time-lapse videos of the painting process from a target painting.  This figure shows 10  keyframes from the generated painting process for two paintings. 
        By training on acrylic paintings with a similar artistic style to that of the first example in this figure, our method is capable of handling a diverse range of styles (e.g., Van Gogh, above bottom). 
        The resulting videos resemble how human artists typically paint, for example, from back to front, focusing on semantic objects or regions at a time, and employing layering techniques.  Images courtesy Catherine Kay Greenup and Rawpixel.
        }
        \label{fig:teaser}
\end{teaserfigure}

%
%

%
%


\maketitle
\setcounter{footnote}{1}

\section{Introduction}
\begin{figure*}[!t]
    \centering
    \includegraphics[scale=0.52]{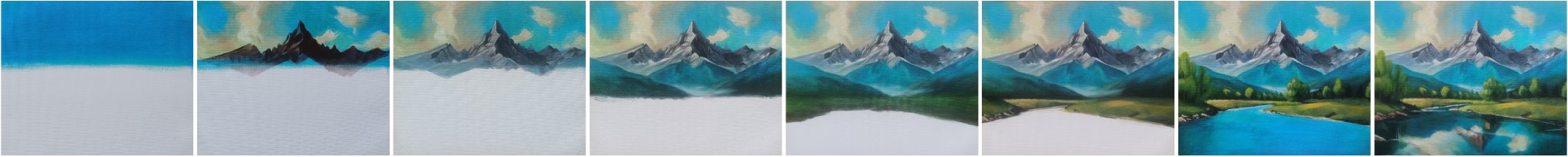}
    \caption{
    \textbf{\bf How a real artist paints.}
    Time lapse from a real painting video, representative of the training painting style. The artist uses a back-to-front order with layering techniques, starting with the sky, then clouds, mountains, and other elements. The artist typically focuses on one semantic region at a time. 
    }
    \label{fig:real_data}
\end{figure*}

When we look at a painting, we see only the final outcome of the artist's creative process.  Leonardo da Vinci worked on the Mona Lisa for 16 years -- it would be fascinating to see a time-lapse video of the Mona Lisa's creation.  
While no such video exists for Leonardo, there are many videos online in which artists have filmed the creation of entire paintings.  
These visualizations are fascinating in the way they show hidden layers, structures, and provide insight into the artistic creation process. 
While such visualizations currently exist for only a tiny subset of paintings in the world, we propose to train machine learning models on such data to predict how a wide variety of other paintings could have been made.  While the resulting videos should not be treated as accurate reconstructions of any specific painting's creation, they are nonetheless intriguing and insightful as {\em plausible} reconstructions, in capturing rules that many painters employ, such as layering, back-to-front-ordering, and focusing on objects/regions in stages~\cite{Dozier_2007,Bob_ross_1987}.
While our approach was trained only on acrylic paintings of landscapes, as shown in Fig.~\ref{fig:real_data}, we believe that future models could produce plausible visualizations for almost any piece of art.

\textcolor{black}{Previous approaches~\cite{huang2019learning,singh2021combining,ganin2018synthesizing,singh2021intelli,de2023segmentation,hu2023stroke,zou2021stylized,liu2021paint} rely on hand-crafted painting principles instead of learning them from the real painting processes.
The most relevant work, Timecraft~\cite{zhao2020painting}, generates time-lapse videos by learning from actual painting videos. However, it operates only on low-resolution (50x50 pixels) patches and therefore lacks holistic semantic context.
}

We define this task as an autoregressive image generation problem. It begins with a blank canvas, which is iteratively updated based on its current state and the target painting. This sequential updating continues until the artwork is completed. To implement this, one possible approach, \textcolor{black}{similar to Timecraft~\cite{zhao2020painting},} is to train a network that takes the current and target images as inputs and outputs the updated canvas at each step. However, we've found that a purely pixel-based approach struggles to produce reasonable results in practice, and thus we incorporate additional semantic analysis.

In particular, we draw inspiration from the techniques used by human artists.  
Consider the second painting in Fig.~\ref{fig:teaser}. Initially, an artist decides on the semantic content to depict -- such as the sky -- and selects appropriate areas of the canvas for this content, such as the upper portion.  Subsequently, the artist applies specific paints to these regions, using blue for the sky, while leaving contents such as stars for later stages.
Building on this observation, we design a two-stage method that decomposes  the instruction generation and canvas rendering. 
In the first stage, we generate a textual instruction and a corresponding region mask from the current and target images.  
The textual instruction provides high-level guidance on the order of painting, and its corresponding region mask directs the focus area.
In the second stage, a diffusion-based renderer is proposed to leverage the textual instruction and region mask, in conjunction with the current and target images, to effectively update the canvas.

Our method can handle in-the-wild landscape paintings across diverse artistic styles (\eg, Impressionism and Realism) and color themes, ranging from dark to bright. We demonstrate that our approach surpasses current state-of-the-art methods in creating high-quality, human-like painting videos, supported by both qualitative and quantitative results, as well as human studies.

\section{Related Work}

\subsection{Painting Process Generation}

\noindent \textbf{Stroke-Based Rendering}.
This is a computer graphics technique that creates non-photorealistic images by placing discrete elements like brush strokes instead of traditional pixels~\cite{haeberli1990paint,hertzmann1998painterly,tang2017animated,frans2018unsupervised,litwinowicz1997processing,ha2017neural,ganin2018synthesizing,xie2013artist,liu2023painterly,jia2019paintbot,fu2011animated}. 
The painting process can be obtained by visualizing the sequence of element placement. However, many studies mainly focus on generating images in various artistic styles~\cite{zou2021stylized,litwinowicz1997processing,xie2013artist,liu2023painterly,kotovenko2021rethinking}. They often overlook the order and position of brush stroke placement, resulting in a non-human-like painting process. 
To produce a more human-like painting process, recent work constrains the placement of brush strokes based on predefined painting principles using techniques such as reinforcement learning (RL)~\cite{huang2019learning,singh2021combining,ganin2018synthesizing,singh2021intelli,de2023segmentation,hu2023stroke,zou2021stylized} or Transformers~\cite{liu2021paint}. 
For example, \cite{liu2021paint} developed a Transformer-based~\cite{vaswani2017attention} feed-forward painter that applies a coarse-to-fine painting strategy. Initially, the painter works from a blurry (downsampled) version of the target painting, progressively refining the canvas using higher-resolution versions until the painting is complete.

However, these techniques face two limitations:
(1)  They rely on hand-crafted painting principles that do not accurately mimic the actual human painting process, whereas our method learns from real-world painting data.
(2) Their parameterized brush strokes fail to capture the full variation observed in real painting processes and  result in an approximate version of the target painting.
In contrast, our diffusion-based renderer effectively handles these variations and recreates the target painting more accurately.

\noindent \textbf{Pixel-Based Generation.}
This stream of work generates the painting process by directly updating pixels on the canvas~\cite{zhao2020painting,leiser2021ai,wang2024intelligent}.  
\cite{wang2024intelligent} used a Vision Transformer (ViT)~\cite{dosovitskiy2020image} to map a target image to a series of 9 intermediate images via a non-autoregressive approach. However, this method is specifically tailored for traditional Chinese painting.
The work most related to ours is Timecraft~\cite{zhao2020painting}, which generates time-lapse videos from paintings by learning from actual painting videos. However, their method is limited to low-resolution (50x50 pixels) crops, neglecting the full artwork's context. Our method handles full paintings with higher resolution.

\subsection{Diffusion Models}
Diffusion models have recently demonstrated their success in various tasks such as text-to-image~\cite{rassin2022dalle,saharia2022photorealistic,Rombach_2022_CVPR}, image-to-image translation~\cite{chen2023total,yang2022paint}, and image-to-video synthesis~\cite{hu2023animate,blattmann2023stable}.  \cite{blattmann2023stable} developed a latent video diffusion model capable of generating videos from an initial frame. 
\cite{hu2023animate} proposed a diffusion framework to synthesize character animation videos from a reference image and a sequence of target poses.  This framework includes a component called ReferenceNet to inject features from the reference image directly  into the denoising UNet of the diffusion model.
In this paper, we adopt the ReferenceNet to inject the target image into our diffusion-based renderer.

\section{Our Method}
\label{sec:method}

In a real painting process, an artist continually applies changes to the current state of the canvas. Our method formulates this process as an autoregressive image generation problem.
Given a target painting $I_T$ as input, we reconstruct a time-lapse video consisting of $T$ \textit{keyframes} $\{\hat{I}_1, ..., \hat{I}_{T}\}$, starting from a blank canvas $I_0$ progressively towards the target $I_T$.
Each \textit{keyframe} transition represents a fixed time interval, a feature of time-lapse videos. 
We start by presenting a one-step canvas rendering approach for each canvas update and discussing its limitations in Sec.~\ref{sec:naive}. 
This motivates our two-stage design, which incorporates additional semantic analysis. The details of the training process are described in Sec.~\ref{sec:cond_signal} and \ref{sec:canvas_rendering}, while the testing is covered in Sec.~\ref{sec:inference_twostage}.
Fig.~\ref{fig:pipeline} shows an overview of the method.

\subsection{One-Stage Canvas Rendering Approach}
\label{sec:naive}
For clarity, we will refer to step $t-1$ as the current step and step $t$ as the next step. The goal is to render next image $\hat{I}_t$ based on $\hat{I}_{t-1}$ and $I_T$. We approach this as an image translation problem, and design a diffusion-based renderer based on the denoising UNet $g_u$ from Stable Diffusion~\cite{Rombach_2022_CVPR}. 
This renderer leverages image priors to enhance image quality by initializing $g_u$ with pretrained weights from Stable Diffusion.

\begin{figure*}[!t]
    \centering
    \includegraphics[scale=0.8]{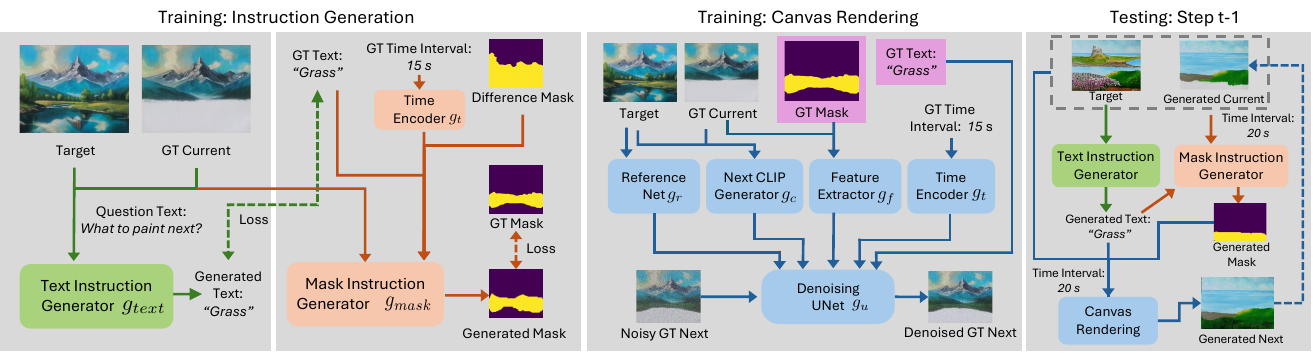}
    \caption{\textbf{Method overview.} The training has two stages. The instruction generation stage (left two gray boxes) includes the text instruction generator (green) and the mask instruction generator (light orange). These generators produce the text and mask instructions essential for updating the canvas in the next stage.
    The second stage is canvas rendering (third gray box), where a diffusion-based renderer generates the next image based on multiple conditional signals, such as text and mask instructions.  
    Omitting the text and mask (purple boxes) yields the one-stage method described in Sec.~\ref{sec:naive}.
    To simplify the figure, we omit the VAE encoder, CLIP encoder, and text encoder. 
 During testing at step $t-1$ (last gray box), we first generate a text instruction (green arrows), which is then used to create a region mask (orange arrows). Both are then provided to the canvas rendering stage to produce the next image (blue arrows).  Image courtesy Catherine Kay Greenup.
    }
    \label{fig:pipeline}
\end{figure*}

We train  this diffusion-based renderer with inputs $\{I_{t-1}, I_{T}, \Delta_t\}$, where $I_{t-1}$ is the ground-truth (GT) current image, and $\Delta_t$ is the actual time interval between $I_{t-1}$ and GT next image $I_t$. 
By incorporating $\Delta_t$, we model the time spent on each update in the painting process, enabling the generation of videos with time-informed keyframes during testing.
The output $\hat{I}_{t}$, which predicts the next image, is supervised using $I_{t}$.  
For supervision, we start with a clean latent $z_0 = E_I(I_t)$, where $E_I$ is the pretrained VAE encoder in Stable Diffusion. We then apply the forward process to produce a noisy latent $z_s = \beta_s z_0 + (1 - \beta_s) \epsilon$, where $s$ denotes a randomly sampled denoising timestep, $\epsilon$ represents Gaussian noise, and $\beta_s$ is a weighting parameter dependent on $s$. The denoising UNet $g_u$ is then employed to denoise $z_s$. We update the parameters of the renderer by minimizing the following loss function:
\begin{equation}
\mathcal{L} = \mathbb E_{s,z_0,\epsilon} || g_u (z_s, c, s) - \epsilon||_2^2,
\label{eq:loss}
\end{equation}
where $c$ represents the conditional signals of $g_u$, consisting of features of $\{I_T, I_{t-1}, \Delta_t\}$.
We extract these features using \textbf{ReferenceNet} $g_r$, \textbf{feature encoder} $g_f$, \textbf{time encoder} $g_t$ and \textbf{next CLIP generator} $g_c$. 
During the training, we jointly update $g_u$, $g_r$, $g_f$, $g_t$, and $g_c$ using Eq.\ref{eq:loss}.
Please refer to the gray box ``Training: Canvas Rendering'' in Fig.~\ref{fig:pipeline} (excluding the mask and text in the purple box).

\noindent\textbf{ReferenceNet $g_r$}.
The target image $I_T$ is integrated into  $g_u$ through  $g_r$, as introduced in~\cite{hu2023animate}. 
The ReferenceNet $g_r$ utilizes the same UNet architecture and  pretrained weights (for initialization) from Stable Diffusion~\cite{Rombach_2022_CVPR}, ensuring feature extraction within the same feature space as $g_u$.
It takes the target image $I_T$ as input, and extracts intermediate feature maps from its self-attention layers.
These feature maps are then fused into $g_u$ by concatenating them with corresponding feature maps from the same layers in $g_u$. 
We opt for the design of ReferenceNet because it fuses features more effectively than other designs, such as ControlNet~\cite{zhang2023adding}, in practice.

\noindent\textbf{Feature encoder $g_f$.}
It consists of a shallow convolutional neural network (CNN) to encode $I_{t-1}$, denoted as $g_f(I_{t-1})$. 
We found that this shallow network effectively learns features of $I_{t-1}$ while saving computational time.
The encoded feature map has the same spatial resolution of  $z_s$ (noisy latent of $I_t$), and is concatenated with $z_s$ along channel dimension as spatial input to the $g_u$ during training. 
The first layer of $g_u$ is modified to adjust to the new channel dimension. 

\noindent\textbf{Time encoder $g_t$ and next CLIP generator $g_c$.}  First, we design a time encoder $g_t$ that applies positional encoding -- same as that used in NeRF~\cite{mildenhall2021nerf} -- to the $\Delta_t$, followed by a multi-layer perceptron (MLP) that maps the positional encoding feature dimension from 21 to 768.
This results in the time embeddings $g_t(\Delta_t)$.
The advantage of positional encoding lies in its ability to adapt to varying $\Delta t$ across different training samples. This continuous encoding enhances interpolation capabilities, useful for handling specific time intervals during testing.
Moreover, to provide additional painting guidance for $g_u$, we introduce the next CLIP generator $g_c$, implemented as an MLP. 
 This module inputs the CLIP embeddings of $I_{t-1}$ and $I_T$, and outputs the predicted CLIP feature of the next image. This predicted CLIP feature and $g_t(\Delta_t)$ are concatenated and fed into the cross-attention layers of $g_u$. 


Fig.~\ref{fig:ablation}~(b) illustrates the results of using this model without incorporating $\Delta_t$. In this scenario, significant content is added in a single update, which does not effectively represent the progressive nature of the painting process. When $\Delta_t$ is included through $g_t$ (Fig.~\ref{fig:ablation}~(c)), the new content volume per update is better controlled; however, this approach still results in unnatural rendering of mountains, as the yellow layers prematurely appear before the completion of the green mountain base. This indicates that a purely pixel-based network struggles to fully capture the semantics of the painting, prompting the need for more explicit semantic controls.

\subsection{Training: Instruction Generation}
\label{sec:cond_signal}
To address the aforementioned issues of the purely pixel-based method, we draw inspiration from typical artistic practices, where artists decide what to paint and where, then apply paint accordingly.
Based on this observation, we propose a two-stage method (Fig.~\ref{fig:pipeline}) where we first generate text and mask instructions as semantic guidance. These instructions are then used to steer the diffusion-based renderer.
In this section, we describe the training of a text and mask generator. These two generators are used to produce two types of instructions for each canvas update: a text instruction $\hat{p}_t$, describes what semantic content to paint, and a region mask $\hat{M}_t$, specifies the focus regions corresponding to the text instruction.

\subsubsection{Text Instruction Generator}
\label{sec:text_instruct_gen}
Like an artist who envisions a target image and compares it to the current state of the work to decide what to paint next, our text instruction generator must discern visual content and generate appropriate textual instructions.
This provides high-level guidance for the painting order.
We implement this generator using the architecture of a large vision-language model LLaVA~\cite{liu2023llava}. LLaVA accepts a single image combined with a question text prompt and produces a response text prompt.
For training, as the generator requires a single image input, we horizontally concatenate the target image, $I_T$, and the ground-truth current image, $I_{t-1}$, to form the input.
The text instruction $\hat{p}_t$ is generated as follows:
\begin{equation}
    \hat{p}_t = g_{text}([I_T, I_{t-1}], p),
    \label{eq:text}
\end{equation}
where $g_{text}$ is the generator, and $[\cdot, \cdot]$ is horizontal concatenation of two images.
$p$ is the question text prompt (details in supplementary).

To enhance the model's performance and reduce training time, we initialize the text generator, $g_{text}$ with pretrained weights of LLaVA 1.5~\cite{liu2023llava}. The text generator is fine-tuned with full supervision of the ground-truth text instructions $p_t$. 
The leftmost gray box in Fig.~\ref{fig:pipeline} visualizes this process, where ``grass'' is generated as the text instruction for the next step. 

\begin{figure}[!t]
\centering
{\includegraphics[width=0.99\linewidth]{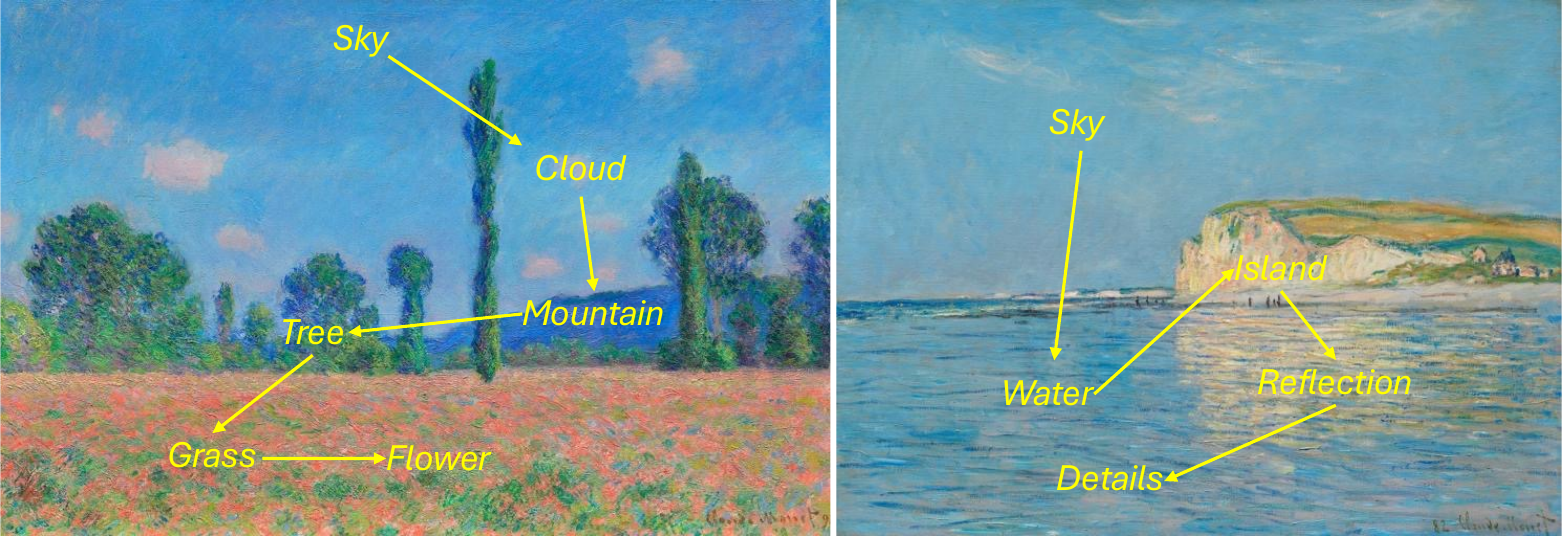}}
\caption{\textbf{Generated text instructions.}
The sequence of generated text instructions (yellow text and arrows) demonstrate a natural painting order, arranging elements from back to front such as clouds over the sky, flowers over grass, and reflections over water. The ``Details'' in the right image refers to water texture and small details on the island. Each text instruction may repeat over multiple frames but is displayed only once to simplify this figure. Images courtesy  the Art Institute of Chicago and Cleveland Museum of Art.
  }
  \label{fig:text_instruction}
\end{figure}

Fig.~\ref{fig:text_instruction} shows the text instructions generated during the painting process of in-the-wild paintings at test time. 
These instructions guide our renderer to use layering techniques, painting from back to front. This demonstrates that $g_{text}$ not only captures the semantic essence of the target paintings but also effectively learns the painting order from the dataset.


\begin{figure*}[!t]
\centering
  \subcaptionbox{Inputs}%
{\includegraphics[width=0.163\linewidth]{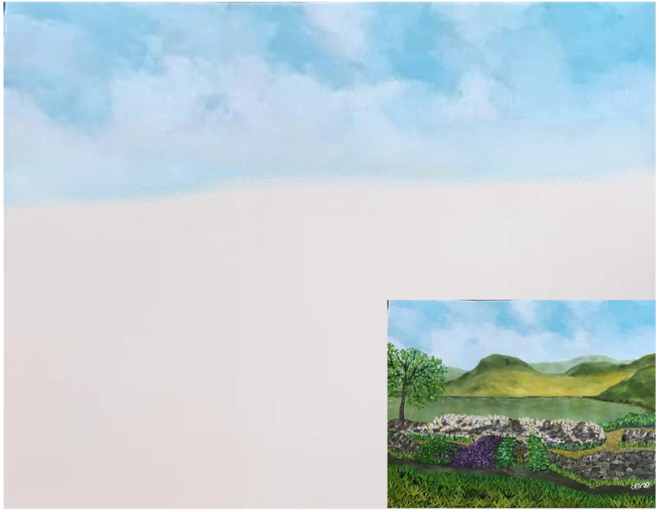}}
%
  \subcaptionbox{CLIP Embed (CE)}
{\includegraphics[width=0.163\linewidth]{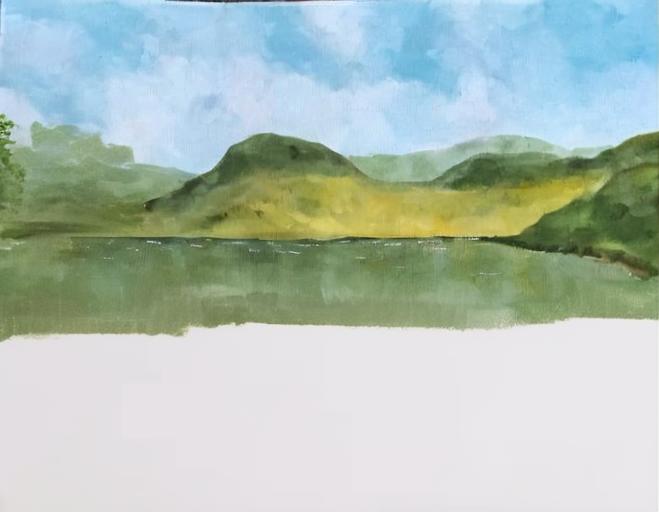}}
%
    \subcaptionbox{CE + Time Embed (TE)}%
{\includegraphics[width=0.163\linewidth]{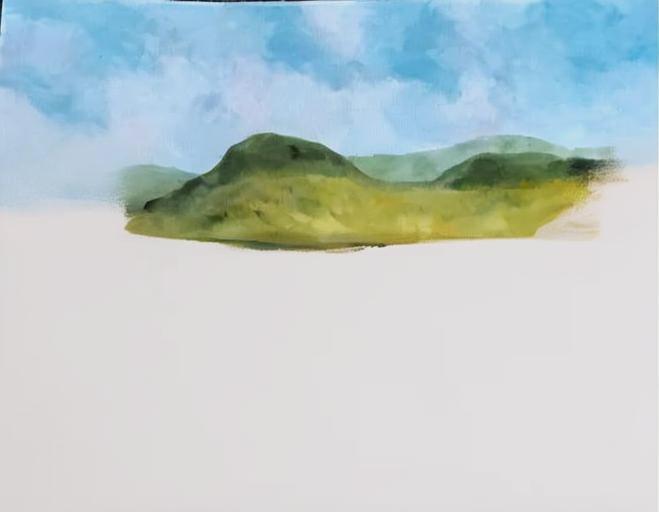}}
    \subcaptionbox{CE + TE + Text}%
{\includegraphics[width=0.163\linewidth]{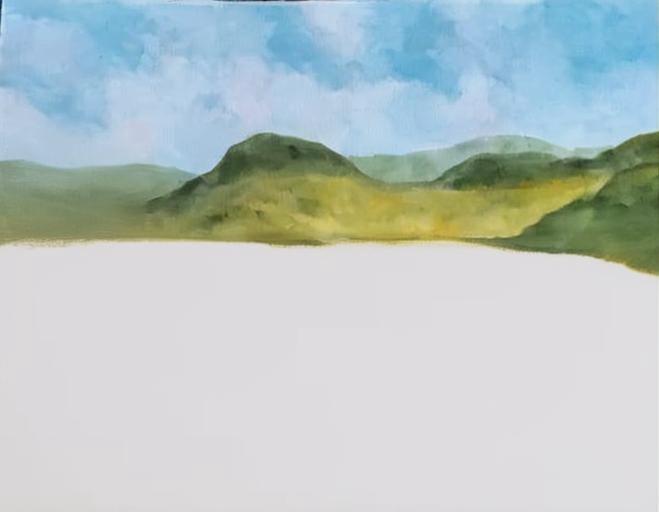}}
    \subcaptionbox{CE + TE + Mask}%
{\includegraphics[width=0.163\linewidth]{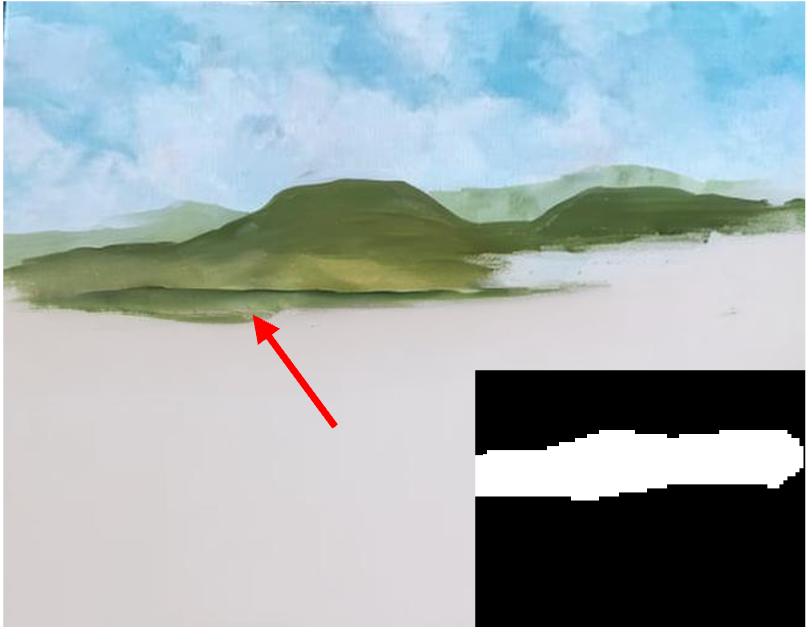}}
    \subcaptionbox{CE + TE + Text + Mask}%
{\includegraphics[width=0.163\linewidth]{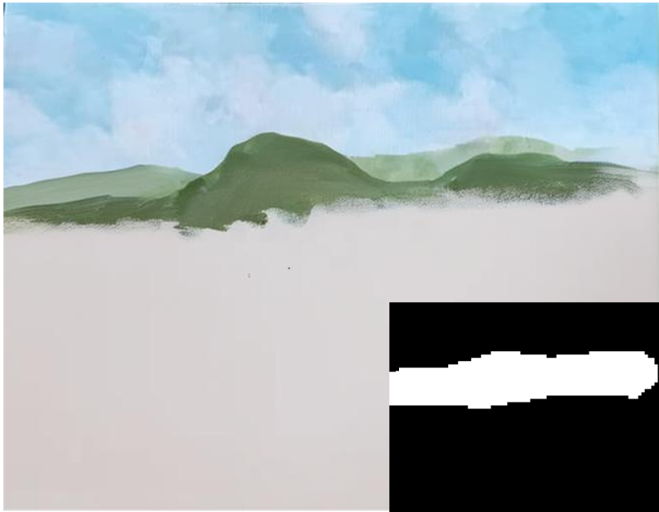}}
\caption{\textbf{Effects of conditional signals.}
(a) shows the current canvas and target image (inset). With only predicted CLIP embeddings of the next image (b), the model generates excessive content per update. Including time intervals (c) properly limits new content volume but results in unnatural mountain rendering. Omitting the mask instruction (d) causes the renderer to complete the mountain area in one step, relying heavily on the text instruction ``mountain''. Omitting the text instruction (e) results in generating some of the green lake (red arrow) before completing the mountain (mask shown in the inset). The full pipeline (f) updates the canvas at a reasonable pace, drawing the top of the mountain in green, before layering on the yellow region.  Image courtesy Catherine Kay Greenup.
}
  \label{fig:ablation}
\end{figure*}

\subsubsection{Mask Instruction Generator} \label{mask_generator}
On top of what to paint, an artist also decides where on the canvas to apply the content. Our method formulates this as a binary region mask, $\hat{M}_t$.  
This mask provides instructions that specify focus regions for each step of the painting process.
For training, as visualized in the second gray box in Fig.~\ref{fig:pipeline}, our  mask instruction generator considers 4 factors to determine the intended painting region:
(1) The GT current image $I_{t-1}$ and target image $I_T$. 
(2) Text instruction $p_t$. During training, we use the ground-truth instruction $p_t$ to ensure that the training is guided by accurate instructional data.
(3) Difference mask $M_d$, which represents areas of the current canvas that are still incomplete relative to the target image. 
This binary mask $M_d$ is derived by computing the difference map between $I_T$ and $I_{t-1}$ using perceptual distance~\cite{zhang2018perceptual}.
This difference map is then binarized using a threshold $\alpha=0.2$, setting pixels with values above $\alpha$ to 1 in $M_d$.
Formally, we define the operations of computing $M_d$ as $D(I_T, I_{t-1}, \alpha)$.
(4) Time interval $\Delta_t$, which can influence the size of the intended painting regions, as less area may be covered when less time is available.

Considering all these factors, we design our mask instruction generator based on the UNet proposed in Stable Diffusion (but without noise as input). 
The cross-attention design of this UNet architecture allows us to seamlessly incorporate textual and time interval information as conditional signals.  


The mask generator $g_{mask}$ has two inputs. 
The first one is spatial input, which includes $I_T$, $I_{t-1}$, and $M_d$.  
Specifically, we begin by encoding $I_T$ and $I_{t-1}$ using the pretrained VAE image encoder $E_I$. This encoder reduces the spatial resolution by a factor of 8 and converts the channel dimension from 3 to 4. We then concatenate these encoded images with the downsampled $M_d$ (notation remains unchanged for simplicity) along the channel axis to form the composite spatial input: $[E_I(I_T), E_I(I_{t-1}), M_d]$. This spatial input is fed into the UNet similarly to Sec.~\ref{sec:naive}.

The second one is the conditional input, which encodes $p_t$ and $\Delta_t$.
We first use the pretrained text encoder $g_p$ in Stable Diffusion to encode $p_t$, producing a text embedding with 77 tokens, each of dimension 768.  
For $\Delta_t$, similar to Sec.~\ref{sec:naive}, we use $g_t$ (same architecture but different weights) to compute time embeddings with dimension 1 x 768.
The time embedding is treated as an additional token and concatenated with the text embeddings to form the 78-token conditional input $[g_p(p_t), g_t(\Delta_t)]$ for cross-attention layers.
Formally, we denote the generation of $\hat{M}_t$ as follows:
\begin{equation}
  \hat{M}_t =  g_{mask} ([E_I(I_T), E_I(I_{t-1}), M_d], [g_p(p_t), g_t(\Delta_t)]).
  \label{eq:mask}
\end{equation}
During the training, we keep $g_p$ and $E_I$ frozen, and update weights in $g_{mask}$ and $g_t$. The training is supervised by the downsampled GT mask $M_t = D(I_t, I_{t-1}, \alpha)$ using the binary cross-entropy loss. 

\begin{figure*}[!t]
    \centering
    \includegraphics[scale=0.72]{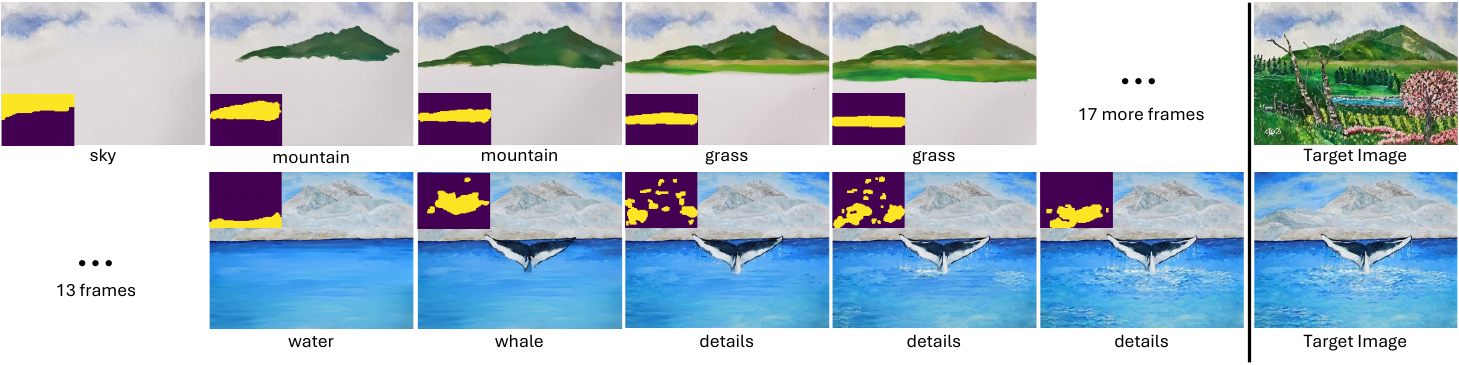}
    \caption{\textbf{Qualitative results with generated texts and masks.}
    We show five consecutive generated keyframes (left) given in-the-wild target images (far-right). The text (caption) and mask (inset) for each keyframe are the generated instructions used to produce that keyframe.
    The first row illustrates the early stages of the process, where our method paints sequentially from back (sky) to front (grass), typically focusing on one subject per keyframe based on the provided text and mask. For instance, keyframes 4 and 5 focus exclusively on grass, disregarding other elements like trees (layered on later) that occupy the same region of the target image.
    The second row depicts the final stages of the painting process for a different painting. Here, our method paints the water before layering on the whale. The process concludes with the addition of final details scattered throughout the painting (as guided by text and mask), mimicking techniques used by human artists.
    The resulting keyframe sequence reflect human-like decisions in painting order, semantic attention, and layering techniques. Images courtesy Catherine Kay Greenup.
    }
    \label{fig:mask_text_vis}
\end{figure*}

\subsection{Training: Canvas Rendering}
\label{sec:canvas_rendering}
Canvas rendering aims to generate $\hat{I}_t$ using the current and target images, alongside text and mask instructions, within a specified time interval.  
We modify the diffusion renderer introduced in Sec.~\ref{sec:naive} to integrate text and mask instructions while keeping other components unchanged, as shown in Fig.~\ref{fig:pipeline} (third gray box). For training, we use the ground-truth text and mask. Specifically, we replace $g_f(I_{t-1})$ with $g_f([I_{t-1}, M_t])$. Here $[\cdot, \cdot]$ refers to concatenation along channel axis. We modify first layer of $g_f$ to handle this change. 
The text $p_t$ is encoded by $g_p$, concatenated with predicted CLIP embeddings and time embeddings, and then input into the cross-attention layers. 
\textcolor{black}{We found these CLIP embeddings provide semantic features beyond what is captured by explicit text and mask instructions alone. }
For instance, consider column 2 in row 1 of Fig.~\ref{fig:mask_text_vis}, where the text ``mountain'' and its corresponding mask serve as the instructions. These instructions do not specify whether the mountain should be painted with intricate details or in a rough, preliminary form, with finer details to be added later. 
\textcolor{black}{Without CLIP embeddings, the model completes the mountain in full detail, deviating from the painting style of the training set. Incorporating the embeddings helps guide the model to adhere to the painting style.
}

Fig.~\ref{fig:ablation} demonstrates the impact of excluding various conditional signals. By omitting the mask, variant (d) generates the entire semantic content (\ie, mountain) at once, underscoring the importance of pixel-level mask guidance. Without text for high-level semantic guidance, variant (e) unexpectedly paints some of the green lake on the canvas. In contrast, the full pipeline (f) achieves the most natural result by integrating all these conditions.

\subsection{Test-Time Generation}
\label{sec:inference_twostage}

At inference time, we begin with a blank (white) canvas $I_0$, and update this canvas autoregressively using our trained pipeline to approximate a target painting $I_T$. We use a fixed time interval $\hat{\Delta}_t$ across all steps, following our definition of keyframes. 
This process is terminated when minimal updates are applied (see supplementary).  
We visualize an update in a specific step $t-1$ in Fig.~\ref{fig:pipeline} (the rightmost gray box).
Given the current image $\hat{I}_{t-1}$ predicted by previous step, we first generate the text instruction $\hat{p}_t$ by employing Eq.\ref{eq:text} and substituting $I_{t-1}$ with $\hat{I}_{t-1}$.
Then we compute the mask $\hat{M}_t$ using Eq.\ref{eq:mask} by replacing $I_{t-1}$ with $\hat{I}_{t-1}$, $M_d$ with $\hat{M}_d = D(I_T, \hat{I}_{t-1}, \alpha)$, $p_t$ with $\hat{p}_t$, and $\Delta_t$ with $\hat{\Delta}_t$. 
Finally, to render $\hat{I}_t$, we start with random noise $z_S$ and perform $S$ denoising steps using $\{\hat{I}_{t-1}, I_T, \hat{p}_t, \hat{M}_t, \hat{\Delta}_t\}$ through diffusion renderer. This results in $z_0$ which is decoded by the VAE decoder from Stable Diffusion, producing $\hat{I}_t$.
  Please refer to the supplementary for details on training and test-time generation, including hyperparameters, execution time, and GT text annotations.

Fig.~\ref{fig:mask_text_vis} visualizes consecutive keyframes of the painting process generated by our method, guided by the  text and mask instructions.  
In the early stages of the process (row 1), the method typically focuses on a single semantic class per update.   
In latter stages of the process (row 2), once the background is completed, the focus switches to completing foreground and refining details.
\textcolor{black}{Here, ``details'' refer to fine elements typically painted in the later stages such as the ocean spray. In the second row of Fig.~\ref{fig:mask_text_vis}, columns 4-6 illustrate the finishing of these elements using the same text instruction “details” but different mask instructions over three frames. }

\section{Experiments}


\noindent\textbf{Datasets}.
We collected a dataset of 294 videos with typical acrylic landscape painting processes. The collection features common themes such as mountains, trees, flowers, and lakes, with paintings representing various times of day and weather conditions. Each video averages 9 minutes in length and is often sped up for more efficient viewing. 
The footage comprehensively captures the complete painting process, including views of the canvas and palette as well as continuous hand and paintbrush movements throughout the video.

For data preprocessing, we employed a pretrained segmentation method~\cite{liu2023grounding} to segment all video frames, cropping them to focus solely on the canvas areas. Further, the cropped frames showing hands, paint strokes, or minimal changes were excluded. After preprocessing, we divided the dataset into 265 training and 29 validation paintings. Both subsets have an average of 27 frames per artwork, with time intervals averaging around 23.6 seconds in training and 22.0 seconds in validation between frames. The training and validation sets have 7261 and 783  pairs. 
\textit{During testing, the time interval is set to 20 seconds unless stated otherwise.}

\begin{figure*}[!t]
    \centering
    \includegraphics[scale=0.75]{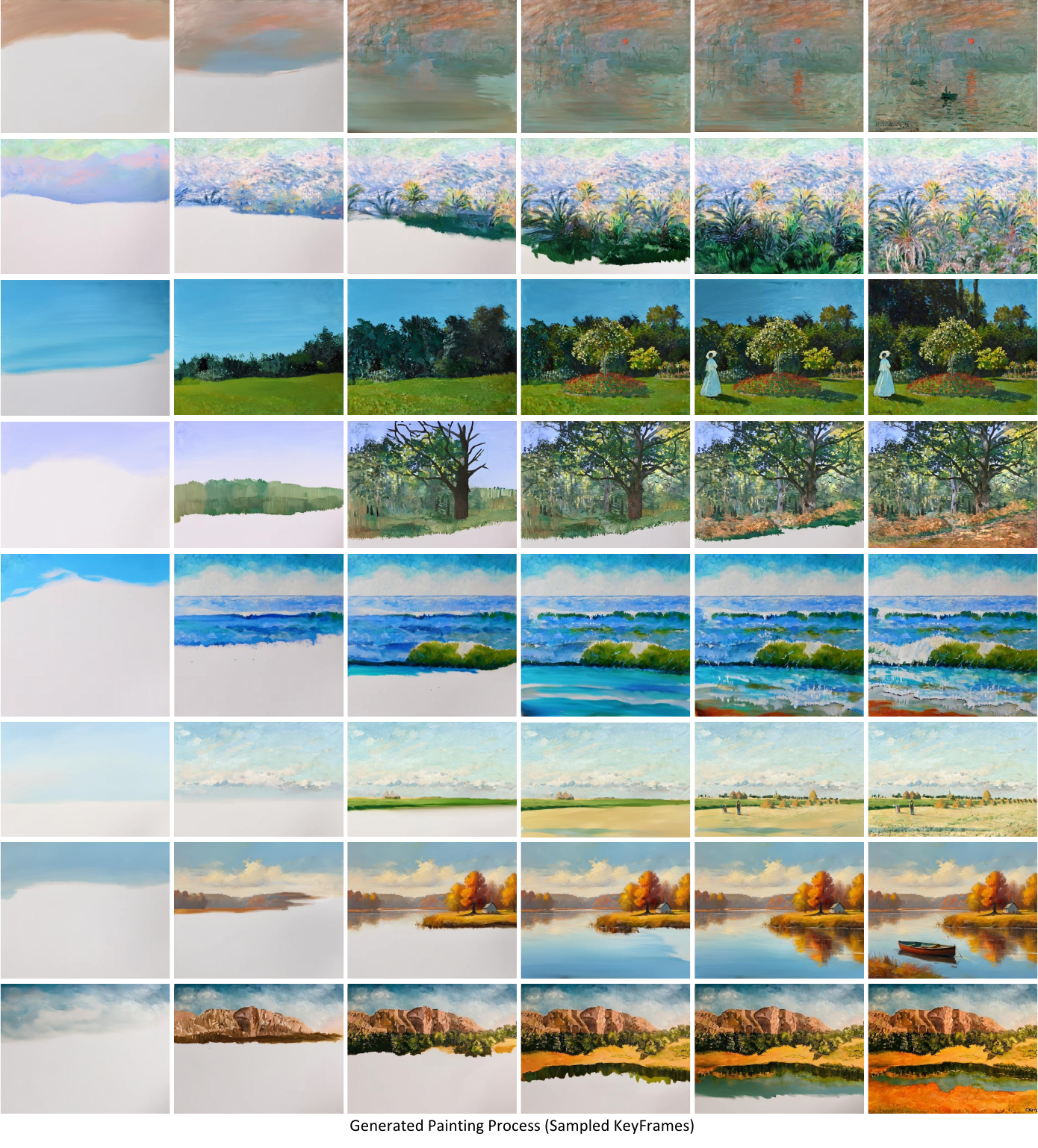}
    \caption{\textbf{Qualitative results.} We show results on in-the-wild paintings (right-most column), where the left five columns are sampled keyframes from the generated painting process. 
Our method effectively handles landscape paintings across various artistic styles, including Impressionism, Post-Impressionism, and Realism. It also adapts to different color themes, as shown. The generated videos showcase human-like layering techniques, painting orders, and semantic region attention.  Images courtesy Catherine Kay Greenup, National Gallery of Art, Washington, and Rawpixel.
        }
    \label{fig:main_results}
\end{figure*}

\noindent\textbf{Our Results}.
Fig.~\ref{fig:main_results} shows the outcomes of our pipeline on in-the-wild paintings. First, the generated painting process resembles human-like painting orderings, typically painting from back to front, saving foreground objects and fine details for the last stages. For example, in row 1, the sequence starts with the sky, moves to the water, and finishes with foreground objects and details such as the boats, the sun, its reflection, and water texture.
Second, each phase of the painting process focuses on specific semantic objects or areas.
For instance, transitions from columns 3 to 6 in row 1 mainly focus on the sun, its reflection on the water, and the boats, respectively.
Third, the underlayering contents are reliably rendered in the intermediate images when applying the layering techniques. 
For example, the base of the mountain is painted in row 2, column 1, with additional details added in column 2. Similarly, clouds are layered across the sky in row 6, columns 1 to 2.
Finally, our method effectively handles artworks of various aspect ratios and artistic styles such as Impressionism. It also accommodates paintings with varied color themes.
 More results are in supplementary.

\noindent\textbf{Baselines}.
We consider three baselines in the main paper (more baselines in supplementary).
(1)\textit{Timecraft}~\cite{zhao2020painting} employs a conditional variational autoencoder to create time-lapse videos from paintings. It trains on real painting videos to emulate the human painting process. However, it handles only low-resolution 50x50 crops from downscaled paintings (126x168) due to computational limits. We trained the model on our dataset following their training strategy. For evaluation, we resized the target painting to 50x50 as input and scale the output videos back to the original painting's resolution. See the supplementary for evaluation on cropped paintings.
(2) \textit{Paint Transformer}~\cite{liu2021paint} generates a stroke-level painting process from an input painting.  This self-supervised method does not rely on real painting videos, but instead employs a hand-crafted coarse-to-fine strategy to mimic human painting process. 
We used the pretrained model provided by the authors for our comparisons. 
Please see supplementary for more stroke-based rendering baselines~\cite{hu2023stroke,zou2021stylized}. 
(3) \textit{Stable Video Diffusion (SVD)}~\cite{blattmann2023stable} produces a 14- or 25-frame video given the first frame as input. We fine-tuned a 14-frame model on our dataset using LoRA\cite{hu2021lora}. During this fine-tuning, we sampled one frame from our training sequences as input and its previous 13 frames as ground truth, padding with white images where necessary. 
For inference, the target painting was input into the fine-tuned model to generate 14 frames. The final frame from this sequence initiated the next set of 14 frames, continuing until a white canvas is achieved.
Please see supplementary for implementation details of all baselines.

\noindent\textbf{Metrics}.
We aim to evaluate the methods in 5 aspects. (1) Human-likeness of the painting order. This measures whether the generated videos mimic human painting order. (2) Focus consistency in canvas updates. This checks if each update targets specific, reasonable areas. (3) Convergence speed towards the target painting. This evaluates the speed at which the generated video progresses towards the target painting.  (4) Adherence to specified time intervals between keyframes. This evaluates whether the temporal progression between generated keyframes aligns with predefined intervals, reflecting the dynamics of an actual painting process. (5) Video quality. 
We design 5 metrics to cover these aspects.  
\begin{itemize}
    \item LPIPS. This addresses aspects (1) and (4). It calculates the perceptual distance between each frame in the real painting video and the closest generated keyframe based on the time. 
    For example, a real frame at 1:17 is compared with the generated frame at 1:20, assuming a 20-second time interval.
    \item IoU (Intersection over Union). It evaluates aspect (2) without considering painting order. We first compute difference masks for consecutive frames in both real and generated videos using the function $D$ (defined in Sec.~\ref{mask_generator}).
For each generated difference mask, we then calculate the IoU with all real difference masks, selecting the highest value as the final IoU.
\item DDC (Difference of Distance Curve). It evaluates aspects (3) and (4) by comparing the distance curves of the generated and real sequences. The distance curve of a sequence depicts its progression towards the target painting, plotting time (x-axis, minutes) against LPIPS distance (y-axis) between target and current images. 
    We use Dynamic Time Warping (DTW)~\cite{muller2007dynamic} to compute the difference between generated and real curves, accommodating time shifts.
    \item DTS (Difference of Time Spent). Evaluating aspect (4), DTS measures the temporal difference in minutes between the durations of the real and generated painting videos.
         \item FID (Fréchet Inception Distance)~\cite{heusel2017gans} compares frames in generated and GT videos for aspect (5).
\end{itemize}

To compute these metrics, for \textit{SVD} and \textit{Timecraft} (trained on our dataset), we set the time interval to 23.6 seconds, matching our training set's average. For the self-supervised \textit{Paint Transformer}, we set it to 2.8 seconds, based on the average training video duration (561 seconds) divided by the number of frames (200).
All metrics are calculated for each painting and averaged across the validation set to derive overall performance metrics.

\begin{figure*}[!t]
    \centering
\includegraphics[scale=0.65]{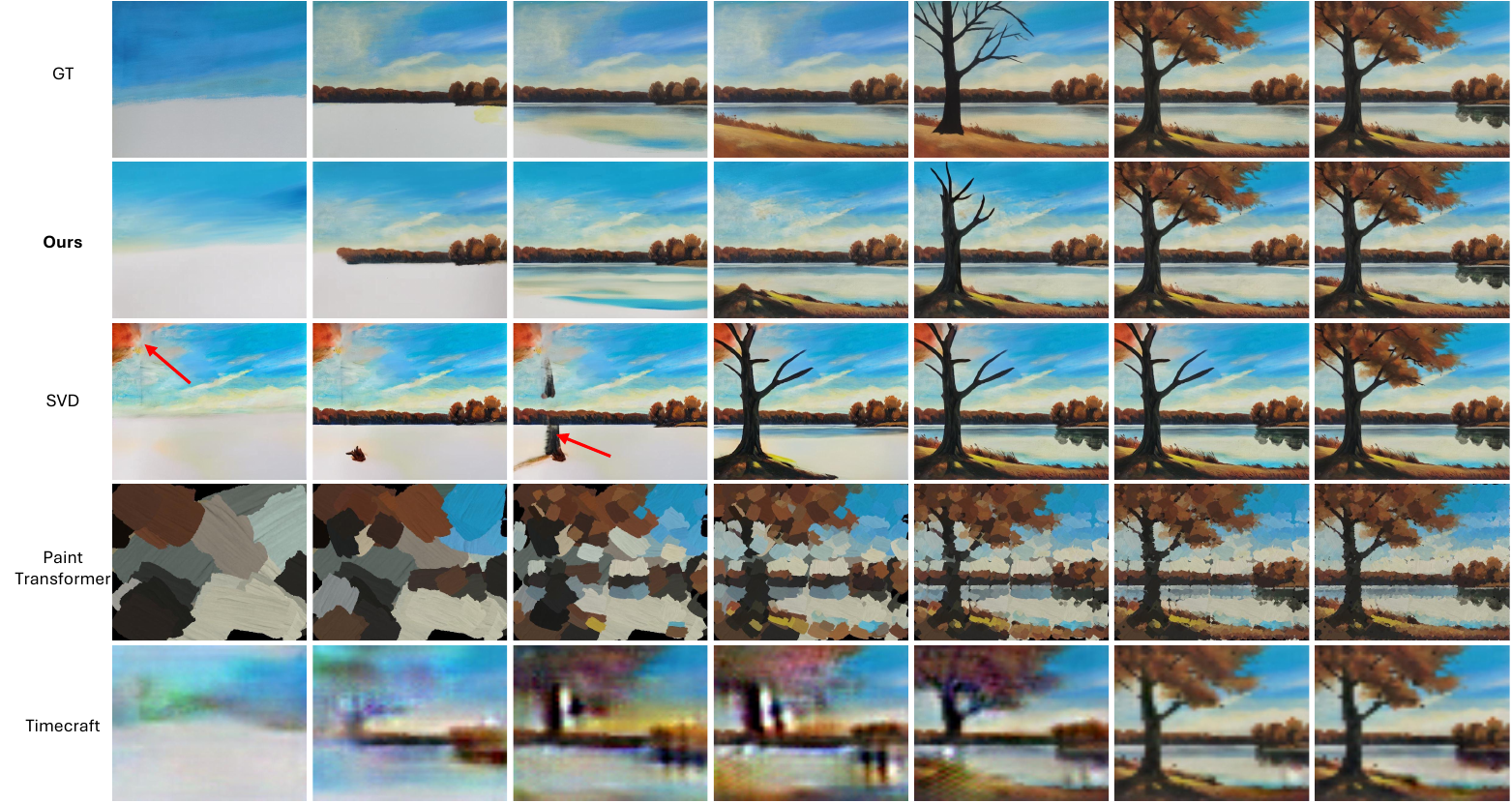}
    \caption{\textbf{Comparison with baselines.}
    Displayed frames are evenly sampled from GT (top row) and generated videos (other rows).
    The last GT frame (top-right) is used as input. \textit{SVD} produces noticeable artifacts such as strange color blocks and truncated tree trunks, as highlighted by the red arrows in columns 1 and 3. Additionally, it produces unreasonable painting orders and tends to get stuck during the process (e.g., minimal change between columns 5 and 6). \textit{Paint Transformer} displays a non-human-like painting order and can only produce a ``stylized'' version of the target painting due to the limitations of parameterized paintstrokes. \textit{Timecraft} results in blurry outputs with noticeable artifacts. In contrast, our method better mimics the human-like painting process and achieves higher visual quality. 
    }
    \label{fig:baseline_comparison}
\end{figure*}

\begin{figure*}[!t]
\centering
  \subcaptionbox{Current Image}%
{\includegraphics[width=0.185\linewidth]{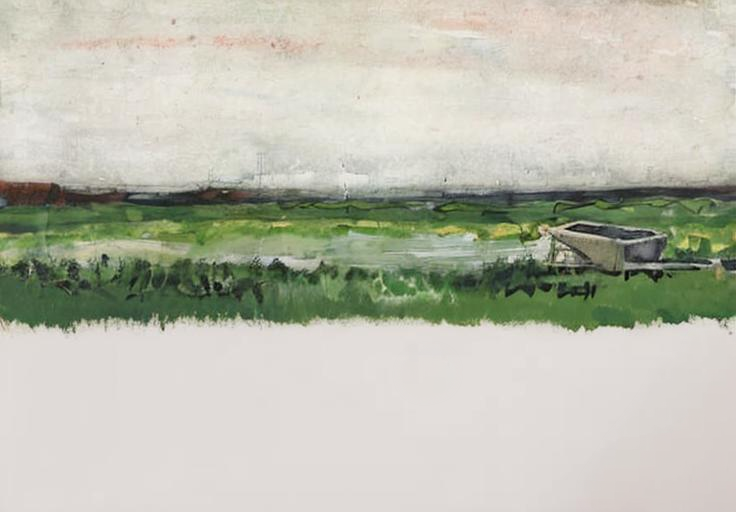}}
  \subcaptionbox{10 seconds}%
{\includegraphics[width=0.185\linewidth]{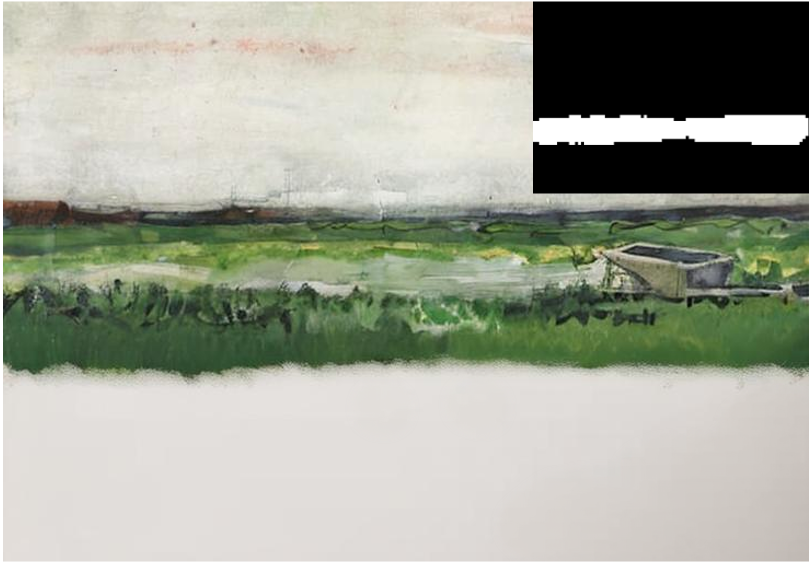}}
  \subcaptionbox{20 seconds}%
{\includegraphics[width=0.185\linewidth]{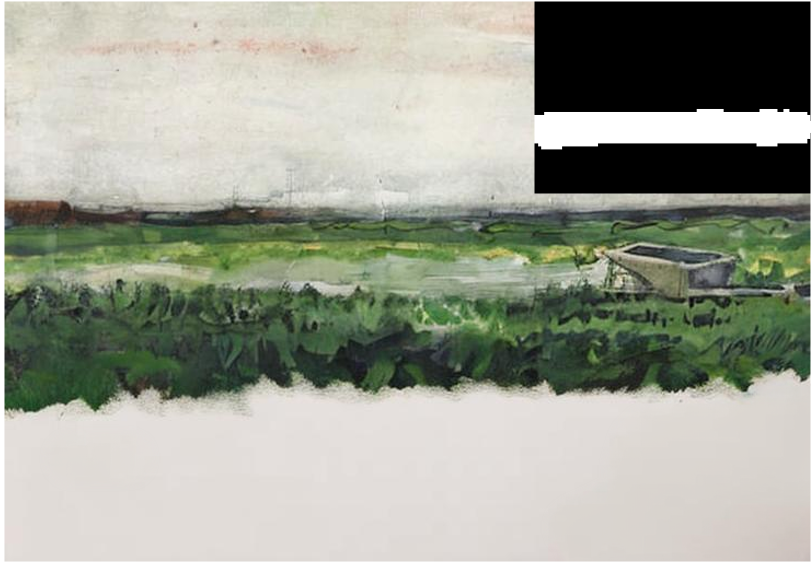}}
  \subcaptionbox{30 seconds}%
{\includegraphics[width=0.185\linewidth]{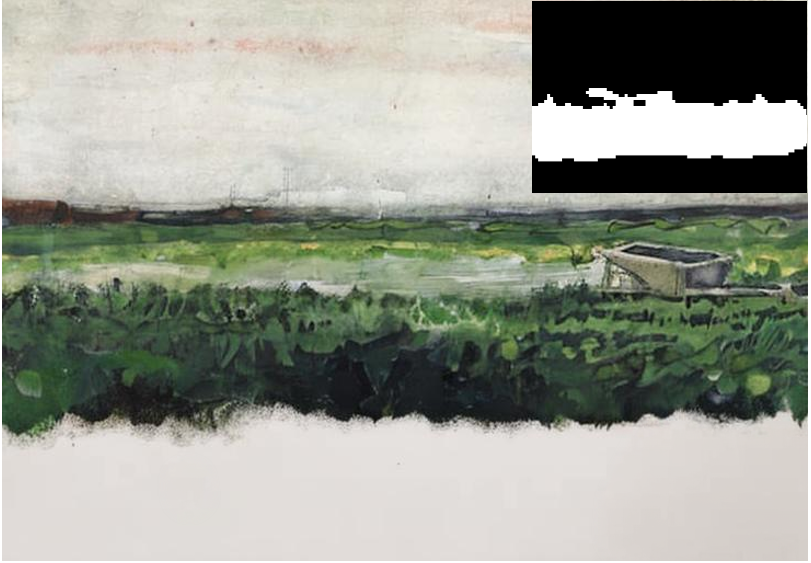}}
\raisebox{-5.5mm}{\rule{1pt}{29.5mm}}%
\hspace{1pt}
  \subcaptionbox{Target Image}%
{\includegraphics[width=0.185\linewidth]{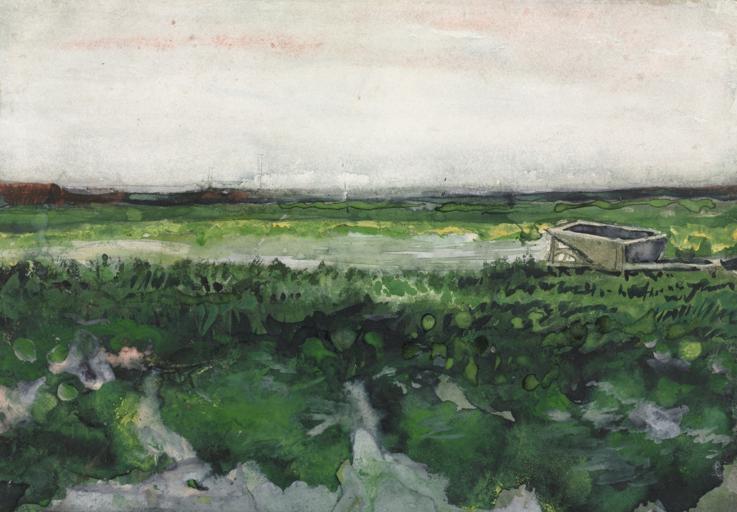}}
\caption{\textbf{Ablation of time interval.} (a) and (e) show the  current  and target image, respectively.  (b) to (d) show the predicted next images and their masks (inset) given different time intervals. As the time interval increases, the size of the predicted mask expands, leading to more extensive updates on the canvas.  Image courtesy Cleveland Museum of Art. 
}
  \label{fig:speed}
\end{figure*}

\begin{table}[!t]
\centering
\resizebox{0.48\textwidth}{!}{
\begin{tabular}{@{} l|ccccc}
\hline
       {Method}  &  {LPIPS} $\downarrow$       & {IoU} $\uparrow$  & {DDC} $\downarrow$   & {DTS} $\downarrow$  
       & {FID} $\downarrow$ 
       \\
\hline
Paint Transformer       &  0.643    &  0.104   & 94.61 &   6.057  &  337.3  \\
Timecraft          &   0.602   &  0.251  &  153.2  & 9.964 & 289.8   \\
SVD          &   0.500   &   0.197  &  135.5  &  8.577 & 168.3    \\
\hline
Ours-TE-TG-MG          &   0.468   &  0.128   &  88.13  &  6.204 &  203.3   \\
 Ours-TG-MG          &    0.447  &  0.139    &  62.79  & 4.913  &  187.4  \\
Ours-TE         &   0.413   &  0.375   & 58.81  &  4.153 & 167.5   \\
Ours-MG         &  0.435    &   0.175  & 61.09  & 3.972  & 182.5 \\
Ours-TG         &   0.399   &  0.400   &  39.41 &  2.120 & 161.1   \\
\textcolor{black}{Ours-RN }        &  \textcolor{black}{ 0.416}   & \textcolor{black}{0.396 }   &  \textcolor{black}{46.71}  &  \textcolor{black}{1.542} &  \textcolor{black}{174.2}\\
Ours-CE         &   0.371   &  0.402    &  34.16  &  1.346 &  158.4  \\
\hline
Ours 10    &    0.369  &  0.349  &  35.27 &  1.693    & 158.7   \\
Ours 30    &     0.387   &  0.353  &  36.26 &  1.933   &  151.7  \\
\hline
Ours          &  \textbf{0.364} &  \textbf{0.418}  & \textbf{32.66} & \textbf{1.273}  &\textbf{150.6} \\
\end{tabular}
}
\caption{\textbf{Comparison with baselines and our ablation variants.} Our full model (with a time interval of 20) outperforms all of them.
}
\label{tab:main_quantitative}
\end{table}

\noindent\textbf{Comparison with Baselines}.
First, we present a qualitative comparison with baselines, as illustrated in Fig.~\ref{fig:baseline_comparison}. 
\textit{SVD} generates artifacts such as the red blocks in the sky regions of column 1. This issue occurs because the target image has a tree that obscures this area, and \textit{SVD} fails to realistically render the obscured regions. It also results in unreasonable painting orders and often stalls during the painting process, leading to extended convergence times (columns 5 and 6).
\textit{Paint Transformer} demonstrates a non-human-like painting order 
\textcolor{black}{in which strokes are added in parallel to different grid cells of a canvas}. \textit{Timecraft} produces visible artifacts in low-resolution videos, likely because its conditional variational autoencoder does not match the image generation quality of diffusion models. Besides, this pure pixel-based method also produces unreasonable painting orders. Our model outperforms these baselines.
Finally, we present the quantitative comparisons in Table.~\ref{tab:main_quantitative}. 
Our pipeline surpasses all tested baselines across all metrics. For additional comparisons and analysis, including the visualization of distance curves and the distribution of their slopes, please refer to the supplementary.


\noindent\textbf{Human Study}.
We conducted a human study on the generated painting processes of 8 in-the-wild paintings. The 33 participants were first presented with 3 examples of the real painting process as reference.  
Then for each painting, the participants were presented with 4 randomly shuffled painting processes (1 for our method and 3 for baselines), and were asked to give a rating from 1 to 5 (higher means better) for each process. We normalized the rating of each example and user to remove the user bias.  Our method achieves the highest average rating, surpassing \textit{SVD} by 1.9 times, \textit{Paint Transformer} by 2.1 times, \textit{Timecraft} by 3.0 times. These show that our method can produce a better painting process than baselines. 
Please see supplementary for more details.

\noindent\textbf{Ablation Study}.
We perform two ablation studies.  The first one evaluates the different components in the pipeline using 7 variants.   
(1) \textit{Ours-TE-TG-MG}: full model excluding time embeddings, text, and mask generators.
(2) \textit{Ours-TG-MG}: full model without text and mask generators. 
(3) \textit{Ours-TE}: full model without time embeddings, omitting the time interval in the pipeline.
(4) \textit{Ours-MG}: full model without the mask  generator.  
(5) \textit{Ours-TG}: full model without the text generator, omitting text in the pipeline.
\textcolor{black}{(6) \textit{Ours-RN}: full model without ReferenceNet, where the target image is inputted to the feature encoder in a manner similar to the current image.}
(7) \textit{Ours-CE}: full model without the next CLIP generator. 
Results shown in Table.~\ref{tab:main_quantitative} and Fig.~\ref{fig:ablation} indicate that the full model surpasses all variants. 
In addition to the discussion in Sec.~\ref{sec:method}, we observe that:
(1) \textit{Ours} outperforms \textit{Ours-TE}, especially on \textit{DDC} and \textit{DTS}, highlighting the importance of time embeddings.
(2) \textit{Ours-MG} exhibits poor performance in IoU, illustrating the effectiveness of mask guidance for focus regions.
(3) \textcolor{black}{\textit{Ours} outperforms \textit{Ours-RN}, showing the effectiveness of using ReferenceNet to inject target image features into the diffusion model. 
}
(4)  \textit{Ours} works better than \textit{Ours-CE}, indicating that the next CLIP generator offers beneficial complementary guidance.

The second ablation study evaluates the effect of different time intervals during testing. We test two variants, \textit{Ours 10} and \textit{Ours 30}, with time intervals set to 10 and 30 seconds respectively. Table.~\ref{tab:main_quantitative} shows that the IoU for these variants is lower than \textit{Ours}, which might be because the average time interval in the validation set is around 22 seconds, closer to \textit{Ours} (20 seconds). This suggests that time embeddings effectively guide the mask generator to produce reasonable size of region masks. 
Additionally, Fig.~\ref{fig:speed} visualizes the impact of different time intervals on a single canvas update. A larger time interval leads to a larger region mask and a more substantial update on the canvas, demonstrating how time intervals can be used to regulate the number of frames required to complete a painting. \textcolor{black}{Different time intervals can also be used at various stages of the painting process based on user needs.}

\noindent\textbf{\textcolor{black}{Analysis of the Text and Mask Generators}}.
First, we compare the entire sequences of our generated text instructions with the sequences of GT instructions in two ways: (a) without considering order, 91\% of our instructions appear in the GT instructions, and (b) taking order into account by computing the longest common subsequence (LCS) between GT and our instructions, 78\% of our instructions are included in the LCS. These show that our text generator provides reasonable instructions for the painting order.

Second, instead of evaluating entire text instruction sequences, we evaluate a single canvas update with GT current image and time interval as input.  This enables the use of GT text and masks for each update during evaluation. Our text generator produces text instructions that align with GT text 72\% of the time, outperforming the pretrained LLaVA model used to initialize it (31\%). When providing the predicted text as input to the mask generator, the predicted mask's IoU is 0.64, slightly lower than using GT text (0.70). These demonstrate the text generator's effectiveness.

Finally, we also evaluate our mask generator based on a single canvas update.
Generating masks by a pretrained segmentation method~\cite{liu2023grounding} using GT text yields an IoU of 0.42, lower than our mask generator's 0.70. Removing the text condition in our mask generator yields a suboptimal IoU of 0.58. These results demonstrate the effectiveness of our mask generator design.

\noindent\textbf{Error Accumulation}.
Our approach of training the diffusion models with GT inputs helps alleviate error accumulation. Specifically, training with GT texts and masks avoids errors propagated from the trained text and mask generators, achieving better LPIPS score of 0.364 compared to training with predicted texts and masks (0.438). 
Additionally, training with GT current frames enhances both stability and efficiency. Furthermore, during inference, using the target image as input helps the convergence of the painting process and further alleviates error accumulation.

\section{Discussions}

\begin{figure}[!t]
    \centering
    \begin{subfigure}[b]{0.48\linewidth}
        \centering
        \includegraphics[width=\linewidth]{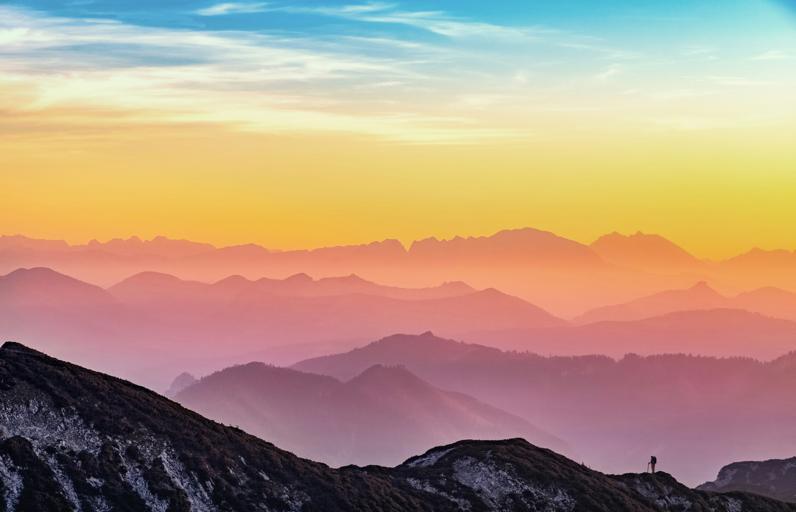}
        \caption{Input}
    \end{subfigure}
    \hfill
    \begin{subfigure}[b]{0.48\linewidth}
        \centering
        \includegraphics[width=\linewidth]{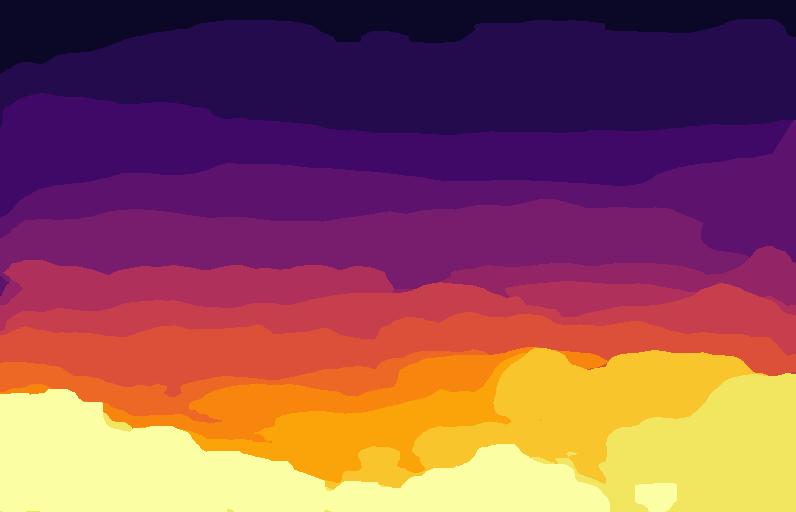}
        \caption{Time Map}
    \end{subfigure}
    \caption{\textbf{Emergent property}. The time map (b), derived from the difference masks, is similar to the (inverse) depth map of the target painting (a). This highlights our method's effectiveness in capturing the typical painting principle of layering from back to front, as learned from the training videos. Image courtesy Simon Berger.}
    \label{fig:emergent_properties}
\end{figure}

\noindent\textbf{Emergent Property}.
Our method showcases an emergent property related to the painting principle, as depicted in Fig.~\ref{fig:emergent_properties}. The time map (b) is derived from difference masks $\hat{M}_t$ between consecutive keyframes in the generated videos. Each mask is weighted by its sequence position $t$ and merged into a single image, with overlapping areas selecting the highest value to form the time map. This map is similar to a depth map, indicating that the back-to-front painting principle -- reflective of the artists' styles in our training dataset -- has been effectively captured by our method.

\begin{figure}[!t]
    \centering
    \includegraphics[scale=0.33]{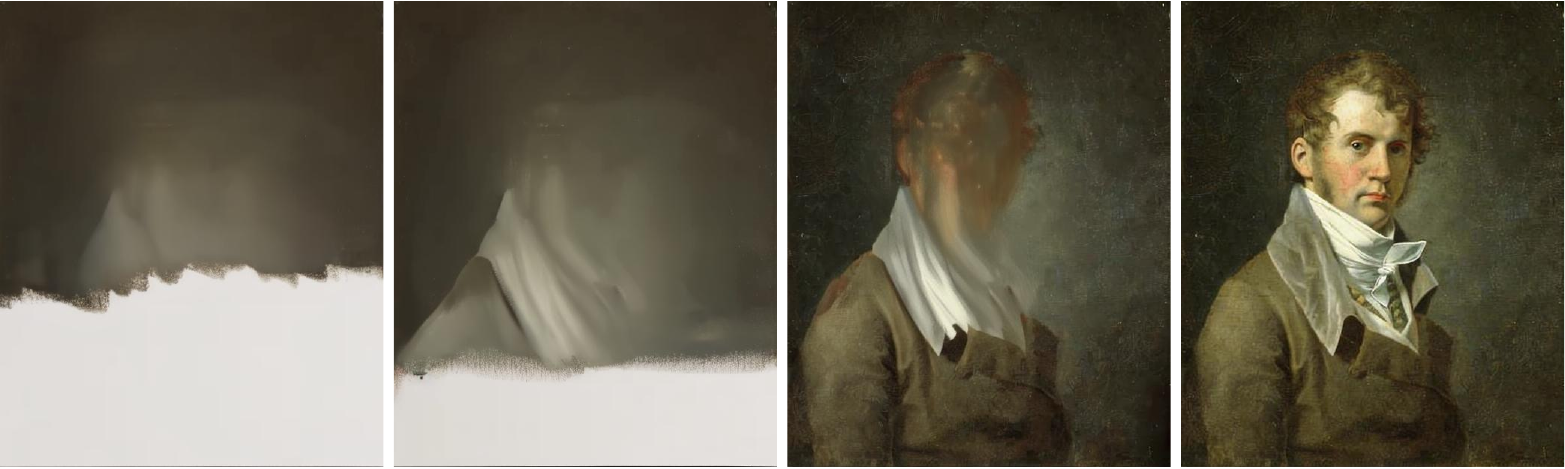}
    \caption{\textbf{Failure cases.} Trained only on landscapes, our method produces unnatural results on portraits. Image courtesy the MET.
    }
    \label{fig:limitations}
\end{figure}

\noindent \textbf{Limitations and Future Work}.
Our method has several limitations.
First, it is trained on landscape paintings and struggles to generalize to other types of paintings like portraits (Fig.~\ref{fig:limitations}). 
Future work will involve extending its applicability to different genres.
\textcolor{black}{Second, our method currently exhibits limited painting styles due to the training data. Training on a larger and more diverse dataset could help the model learn various painting styles. }
Third, our method fails when encountering paintings with large objects (\eg, colosseum) that are uncommon in the landscape. 
\textcolor{black}{
In such cases, the model can still paint the objects with the help of predicted CLIP embeddings, but in an incorrect painting order.
}
Incorporating a semantic map could potentially address this issue.

\begin{acks}
This work was supported by the UW Reality Lab and Google. 
\end{acks}

\newpage
\bibliographystyle{ACM-Reference-Format}
\bibliography{sample-bibliography}




\end{document}



\title{Inverse Painting:  Reconstructing The Painting Process}


\author{Bowei Chen}
\orcid{0000-0002-2225-8796}
\affiliation{%
 \institution{University of Washington}
 \streetaddress{1410 NE Campus Pkwy}
 \city{Seattle}
 \state{WA}
 \postcode{98195}
 \country{USA}}
\email{boweiche@cs.washington.edu}
\author{Yifan Wang}
\orcid{0000-0002-8246-2254}
\affiliation{%
 \institution{University of Washington}
 \city{Seattle}
 \country{USA}
}
\email{yifan1@cs.washington.edu}
\author{Brian Curless}
\orcid{0000-0002-0095-5400}
\affiliation{%
 \institution{University of Washington}
 \city{Seattle}
 \country{USA}
}
\email{curless@cs.washington.edu}
\author{Ira Kemelmacher-Shlizerman}
\orcid{0009-0003-9498-584X}
\affiliation{%
 \institution{University of Washington}
 \city{Seattle}
 \country{USA}
}
\email{kemelmi@cs.washington.edu}
\author{Steven M. Seitz}
\orcid{0009-0000-4214-4078}
\affiliation{%
 \institution{University of Washington}
 \city{Seattle}
 \country{USA}
}
\email{seitz@cs.washington.edu}

\renewcommand\shortauthors{Chen, et al}


\maketitle
\appendix

\section{Implementation Details}
In this section, we present the implementation details. 
\subsection{One-Step Canvas Rendering Approach}
Please refer to Sec.~\ref{sec:renderer} for implementation details.

\subsection{Training: Instruction Generation}
\label{sec:instruct_gen}

For text instruction generator $g_{text}$, we set the question prompt $p$ to: ``There are two images side by side. The left image is an intermediate stage in a painting process of the right image. Please tell me what content should be painted next? The answer should be less than 2 words.''

We obtain the ground-truth text instructions $p_t$ with the assistance of the pretrained LLaVA 1.5 model \textcolor{black}{``LLaVA-v1.5-7B\footnote{https://huggingface.co/liuhaotian/llava-v1.5-7b}''}. Specifically, we horizontally concatenate the ground-truth current image $I_{t-1}$ with the ground-truth next image $I_{t}$. We then input this concatenated image alongside the modified question prompt $p'$, which reads: ``There are two images side by side. The right image is the next step of the left image in the painting process of a painting. Please tell me what is added to right image? The answer should be less than 2 words.'' 
The LLaVA model then outputs the text instructions $p_t$, where we manually correct any inaccuracies in $p_t$.

For fine-tuning $g_{text}$, we employ LoRA and train for 10 epochs, using a learning rate of 1e-4 and a batch size of 16. The fine-tuning process takes approximately 5 hours on a single NVIDIA A100 GPU.

For the mask generator $g_{mask}$, we implement this as the  UNet architecture proposed by Stable Diffusion~\cite{Rombach_2022_CVPR}.  The input layer of the UNet is modified to accept a 9-channel input, tailored for the spatial input components $[E_I(I_T), E_I(I_{t-1}), M_d]$.  
For the time encoder $g_t$, we implement 3 fully connected layers that progressively increase the input feature dimension from 21 to 256, 512, and finally 768. A ReLU activation function follows each fully connected layer, with the exception of the last one to allow for linear output transformation.
We train $g_{mask}$ and $g_t$ for 80k steps with a learning rate of 1e-5 and a batch size of 1. The training process takes approximately 13 hours on a single NVIDIA A100 GPU.

\subsection{Training: Canvas Rendering}
\label{sec:renderer}
The feature extractor $g_f$ takes as input the ground-truth current image $I_{t-1}$ and region mask $M_t$, outputs an encoded feature of them. This feature extractor is implemented as a shallow network containing 9 convolutional layers, which scales the input spatial resolution by 8.  
The next CLIP generator, denoted as $g_c$, accepts the CLIP embeddings of both the ground-truth current image, $CLIP(I_{t-1})$, and the target image, $CLIP(I_T)$, as inputs. It then outputs a prediction for the CLIP embedding of the next image. Within $g_c$, the embeddings $CLIP(I_{t-1})$ and $CLIP(I_T)$ are concatenated along the feature dimension. This concatenated vector is subsequently processed through a three-layer multi-layer perceptron (MLP). The MLP's layers map the features from dimensions of 1536 to 768, 384, and back to 768 respectively. The ReLU activation function is employed at each layer except the final one, where no activation is applied. The time encoder $g_t$ consists of 3 fully connected layers, same as that introduced in Sec.\ref{sec:instruct_gen}. 
For more details on the implementation of ReferenceNet $g_r$, please refer to \cite{hu2023animate}.

We initialize $g_u$ using RealisticVision V5.1~\cite{Civitairv51}.
The models within canvas rendering, namely $g_u$, $g_r$, $g_f$, $g_t$, and $g_c$, are jointly trained using a learning rate of $1 \times 10^{-5}$ and a batch size of 1. Each conditional signal is dropped (set to zero) 10 percent of the time, in accordance with the classifier-free guidance method~\cite{ho2022classifier}. This enables us to control the strength of each conditional signal at the test time inference.  We train the models in 200k steps, taking around 34 hours on a single NVIDIA A100 GPU.  

For the one-stage approach outlined in Sec. 3.1 of the main paper, we use the same training strategy but exclude text and mask instructions.

\subsection{Test-Time Generation}
At a specific step $t-1$, we render the subsequent image $\hat{I}_t$ using the trained pipeline. The denoising process of the diffusion renderer employs a scheduler based on ancestral sampling, specifically utilizing the Euler method steps~\cite{karras2022elucidating}. The denoising timestep $S$ is set to 25. For classifier-free guidance, we assign guidance scales of 5 for text, mask, and time interval, and a scale of 2 for the next CLIP embeddings. Each update in this process takes approximately 4 seconds on a single NVIDIA A100 GPU. The generation process is halted if the perceptual distances between $\hat{I}_{t-2}$ and $\hat{I}_{t-1}$, and between $\hat{I}_{t-1}$ and $\hat{I}_t$, are both less than $1 \times 10^{-3}$.

\subsection{Baselines Details}
For \textit{Timecraft}, we train the model on our dataset using the default settings provided in the official code. The training consists of two stages: pairwise optimization and sequence optimization. In the first stage, we train the model for 500K steps, which takes approximately 4 hours on 2 TITAN XP GPUs. In the second stage, we train the model for 78K steps, which takes around 25 hours on 2 TITAN XP GPUs. We observed that training with more steps will degrade the model performance. 

For \textit{Stable Video Diffusion (SVD)}, we fine-tune a 14-frame model on our dataset using LoRA \cite{hu2021lora}. During fine-tuning, we sample one frame from our training sequences as input and use its previous 13 frames as ground truth, padding with white images when necessary. 
The target image is used as the input frame 40\% of the time, while other images are randomly selected otherwise. We fine-tune the model for 2K steps, which takes around 1.5 hours on 4 NVIDIA A100 GPUs. Fine-tuning for 2K steps yields the best performance; more steps cause the model to produce painting videos that get stuck, while fewer steps result in underfitting, causing the camera viewpoint to shift.

For \textit{Paint Transformer}, we use the pretrained model provided by the authors for our comparisons. This model generates 200 frames given an input painting.
We additionally evaluate two stroke-based rendering baselines~\cite{zou2021stylized,hu2023stroke} \textcolor{black}{and an amodal segmentation baseline~\cite{ozguroglu2024pix2gestalt}} in Sec.~\ref{sec:supp_exp}, please refer to Sec.~\ref{sec:more_baselines} for their implementation details. 

\section{Experiments}
\label{sec:supp_exp}

\subsection{More Results}
 
\begin{figure*}[!t]
    \centering
    \includegraphics[scale=0.75]{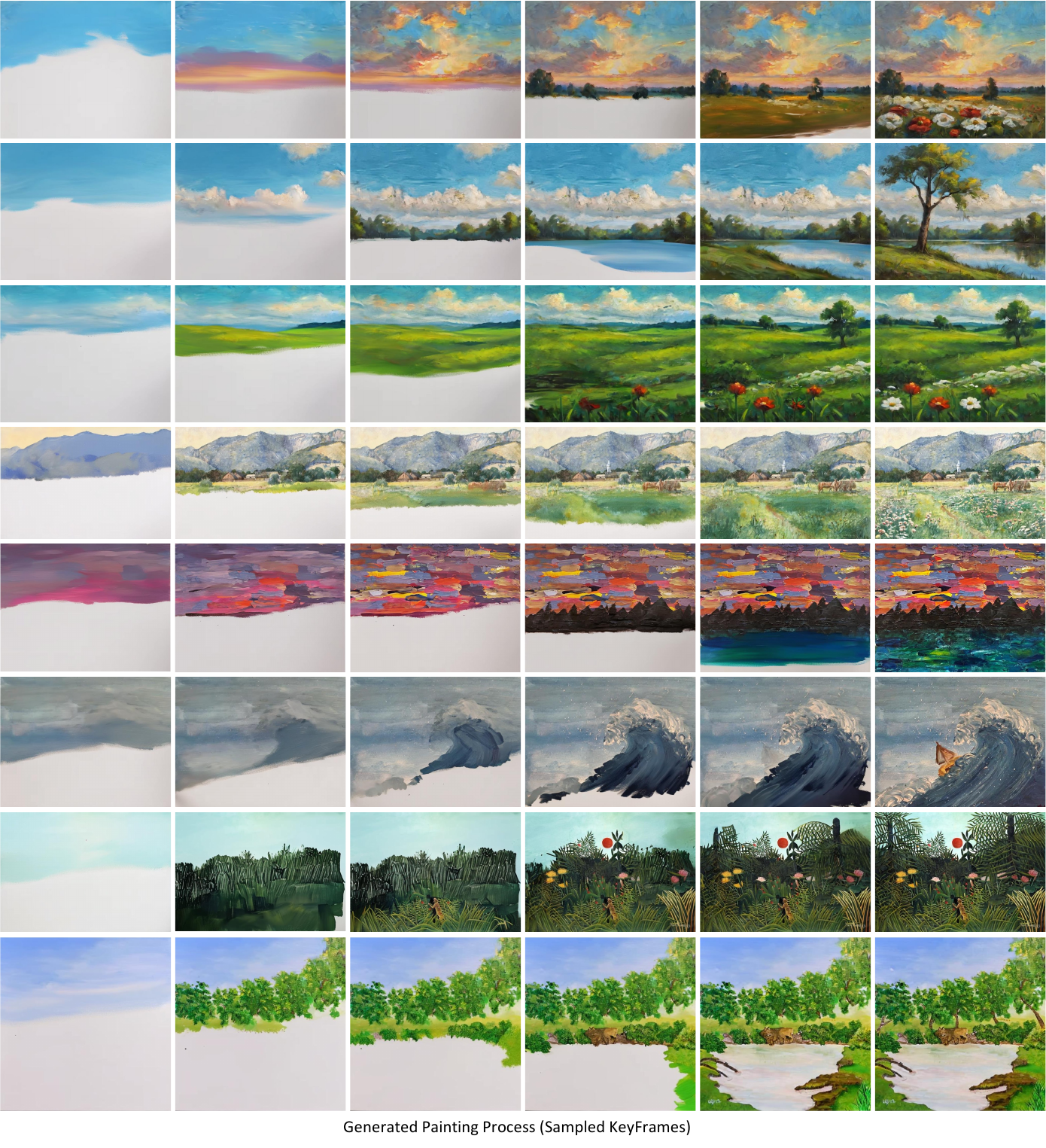}
    \caption{\textbf{More qualitative results on our method.} We show our results on in-the-wild paintings, where the left 5  columns are sampled frames from the generated painting process, and the rightmost column is the target image. 
Our method effectively handles paintings across various artistic styles, color themes, and aspect ratios. The generated sequences showcase human-like painting orders, maintain reasonable focal regions during different phases, and employ layering painting techniques. Images courtesy  Michelle Shlizerman and Catherine Kay Greenup. Image courtesy Julius Zorkoczy, Landscape with a Blooming Meadow, Slovenska narodna galeria, SNG, \url{https://www.webumenia.sk/dielo/SVK:SNG.K_1710}. Image courtesy Henri Rousseau, Virgin Forest with Sunset, Kunstmuseum Basel Museum. 
        }
    \label{fig:supp_main_results}
\end{figure*}

\begin{figure*}[!t]
    \centering
    \includegraphics[scale=0.75]{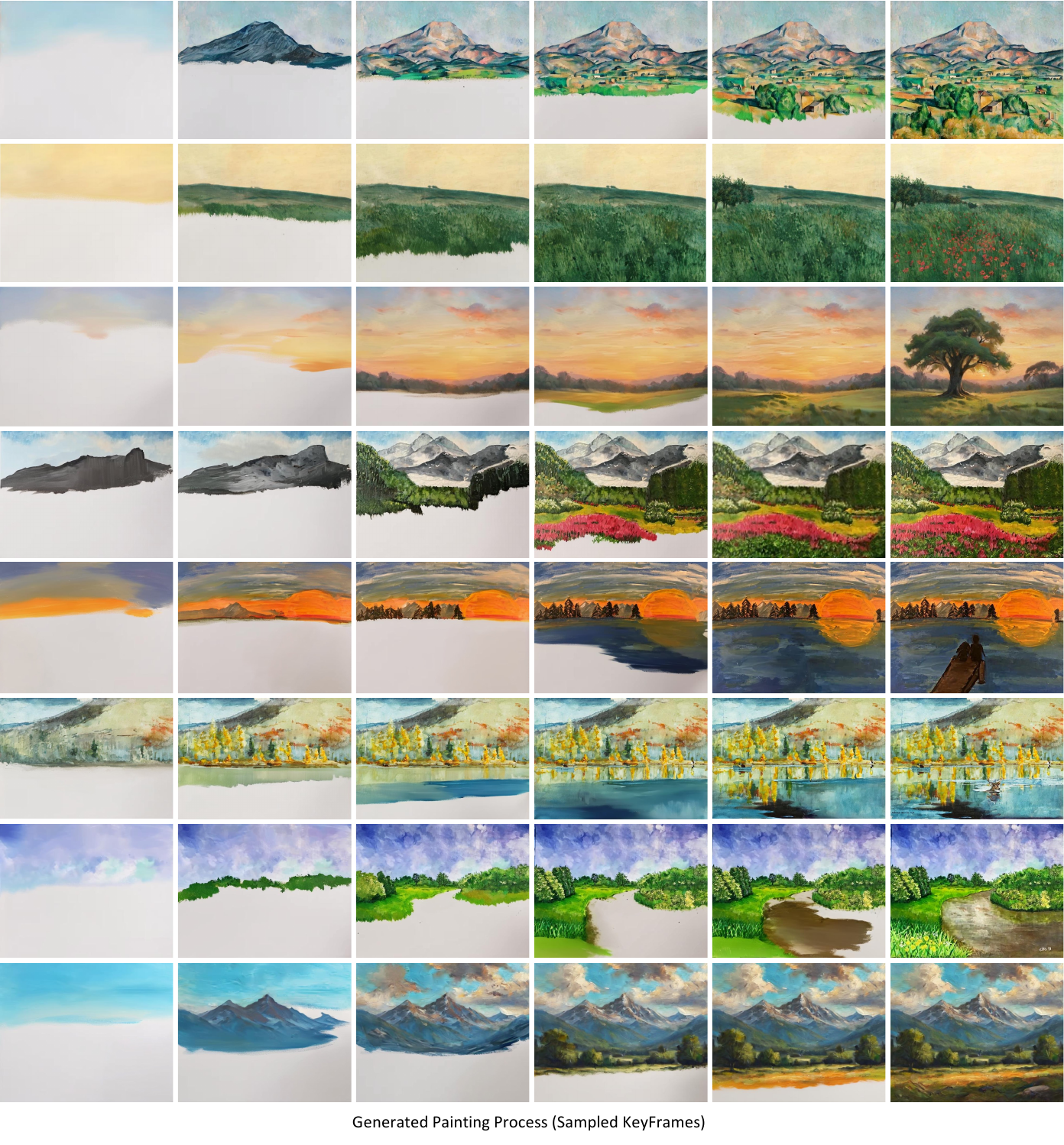}
    \caption{\textbf{More qualitative results on our method.} We show our results on in-the-wild paintings, where the left 5  columns are sampled frames from the generated painting process, and the rightmost column is the target image. Our method effectively handles paintings across various artistic styles, color themes, and aspect ratios. The generated sequences showcase human-like painting orders, maintain reasonable focal regions during different phases, and employ layering painting techniques. Images courtesy Barnes Foundation, Catherine Kay Greenup, The Clark Art Institute and Michelle Shlizerman. Image courtesy František Kaván – Red Poppies, 1910, Slovenská národná galéria, SNG,  \url{https://www.webumenia.sk/dielo/SVK:SNG.O_4549}. 
        }
    \label{fig:supp_main_results2}
\end{figure*}

In Fig.~\ref{fig:supp_main_results} and Fig.~\ref{fig:supp_main_results2}, we show more results of our method. As discussed in the main paper, our method can  handle paintings with different styles and generate human-like painting process in terms of painting order, focal region and layering techniques.


\subsection{More Baselines}
\label{sec:more_baselines}
We compare our method with two additional stroke-based rendering baselines \textcolor{black}{and an amodal segmentation baseline.}

\textit{Stylized Neural Painting}~\cite{zou2021stylized} employs an optimization-based approach featuring a novel neural renderer that mimics vector renderer behavior. Here, the stroke prediction is framed as a parameter search process aiming to maximize the similarity between the input image and the rendered output. We utilize the pretrained network ``the oil-paint brush'' provided by the authors for comparison. This method generates 499 frames given a target painting.

\textit{Compositional Neural Painter}~\cite{hu2023stroke} utilizes a phased RL strategy for predicting paint regions and a painter network to determine stroke parameters. A neural stroke renderer is then trained to apply the strokes onto the canvas based on the predicted stroke parameters. We use the authors' pretrained networks for comparison. This method generates 50 frames from a target painting.

\textcolor{black}{
\textit{pix2gestalt}~\cite{ozguroglu2024pix2gestalt} completes a partially visible object in the image given the partial segment mask of the object.  
We adapt it to out task as follows: (1) segment the target image using \cite{liu2023grounding}, (2) complete each segment using the pretrained model of \textit{pix2gestalt}, (3) place completed segments on the canvas by depth (farther first). The depth of each segment is determined by the average depth of its pixels, where the depth is estimated using a pretrained depth estimation model~\cite{depthanything}.
Please note that we define the painting order heuristically, as the baseline does not support learning this order.
}

Similar to the strategy used for \textit{Paint Transformer} in the main paper, we set the time intervals for these three baselines based on the average training video duration (561 seconds) divided by the number of frames. This results in time intervals of 1.12 seconds for \textit{Stylized Neural Painting} and 11.22 seconds for \textit{Compositional Neural Painter}.
\textcolor{black}{
The time interval of \textit{pix2gestalt} varies for different target images, depending on the number of detected segments in the target image. 
}

\subsection{More Metrics}
We also evaluate the quality of the generated videos using the Fréchet Video Distance (FVD)~\cite{ge2024content}. While FVD might not be ideally suited for assessing time-lapse painting process videos, we include it for the sake of comprehensiveness.

\subsection{Baseline Comparison}

\begin{table}[!t]
\centering
\resizebox{0.48\textwidth}{!}{
\begin{tabular}{@{} l|cccccc}
\toprule
\multicolumn{7}{c}{Evaluation on Full Paintings} \\
\hline
       {Method}  &  {LPIPS} $\downarrow$       & {IoU} $\uparrow$  & {DDC} $\downarrow$   & {DTS} $\downarrow$  & {FID} $\downarrow$ & {FVD} $\downarrow$ \\
\hline
Stylized Neural Painting          & 0.669 & 0.031  &  122.2  &  8.782   & 358.3 & 1457 \\
Compositional Neural Painter          & 0.680 & 0.049  &  91.93 &  7.507   &  374.8 & 1505 \\
\textcolor{black}{pix2gestalt}          & \textcolor{black}{0.609} & \textcolor{black}{0.214}  &  \textcolor{black}{126.3} & \textcolor{black}{9.089}  &  \textcolor{black}{341.9} &\textcolor{black}{1576} \\
Paint Transformer       &  0.643    &  0.104   & 94.61 &   6.057  &  337.3 & 1616 \\
Timecraft          &   0.602   &  0.251  &  153.2  & 9.964 & 289.8  &  1582 \\
SVD          &   0.500   &   0.197  &  135.5  &  8.577 & 168.3   &  1594 \\
\midrule
Ours-TE-TG-MG          &   0.468   &  0.128   &  88.13  &  6.204 &  203.3  & 1591 \\
 Ours-TG-MG          &    0.447  &  0.139    &  62.79  & 4.913  &  187.4  &  1468\\
Ours-TE         &   0.413   &  0.375   & 58.81  &  4.153 & 167.5  & 1319 \\
Ours-MG         &  0.435    &   0.175  & 61.09  & 3.972  & 182.5 & 1471\\
Ours-TG         &   0.399   &  0.400   &  39.41 &  2.120 & 161.1   & 1418\\
\textcolor{black}{Ours-RN }        &  \textcolor{black}{ 0.416}   & \textcolor{black}{0.396 }   &  \textcolor{black}{46.71}  &  \textcolor{black}{1.542} &  \textcolor{black}{174.2} &  \textcolor{black}{1464}\\
Ours-CE         &   0.371   &  0.402    &  34.16  &  1.346 &  158.4 & 1326 \\
\midrule
Ours 10    &    0.369  &  0.349  &  35.27 &  1.693    & 158.7  & 1347 \\
Ours 30    &     0.387   &  0.353  &  36.26 &  1.933   &  151.7  & 1279\\
\midrule
Ours          &  \textbf{0.364} &  \textbf{0.418}  & \textbf{32.66} & \textbf{1.273}  & \textbf{150.6}  & \textbf{1273}\\
\midrule
\multicolumn{7}{c}{Evaluation on Cropped Paintings} \\
\midrule
Timecraft       &   0.647   &   0.165  & 166.29  & 6.743  &  363.0 & 1627 \\
Ours          &  \textbf{0.452} &  \textbf{0.296} &    \textbf{56.62} & \textbf{2.545}  &  \textbf{197.2}   & \textbf{1034}\\
\end{tabular}
}
\caption{\textbf{Comparison with baselines and our ablation variants on the full and cropped paintings.} Our full model (with a time interval of 20) outperforms all of them.
}
\label{tab:crop_quantitative}
\end{table}

We present more comparisons with baseline methods on both full and cropped paintings.

\begin{figure*}[!t]
    \centering
\includegraphics[scale=0.72 ]{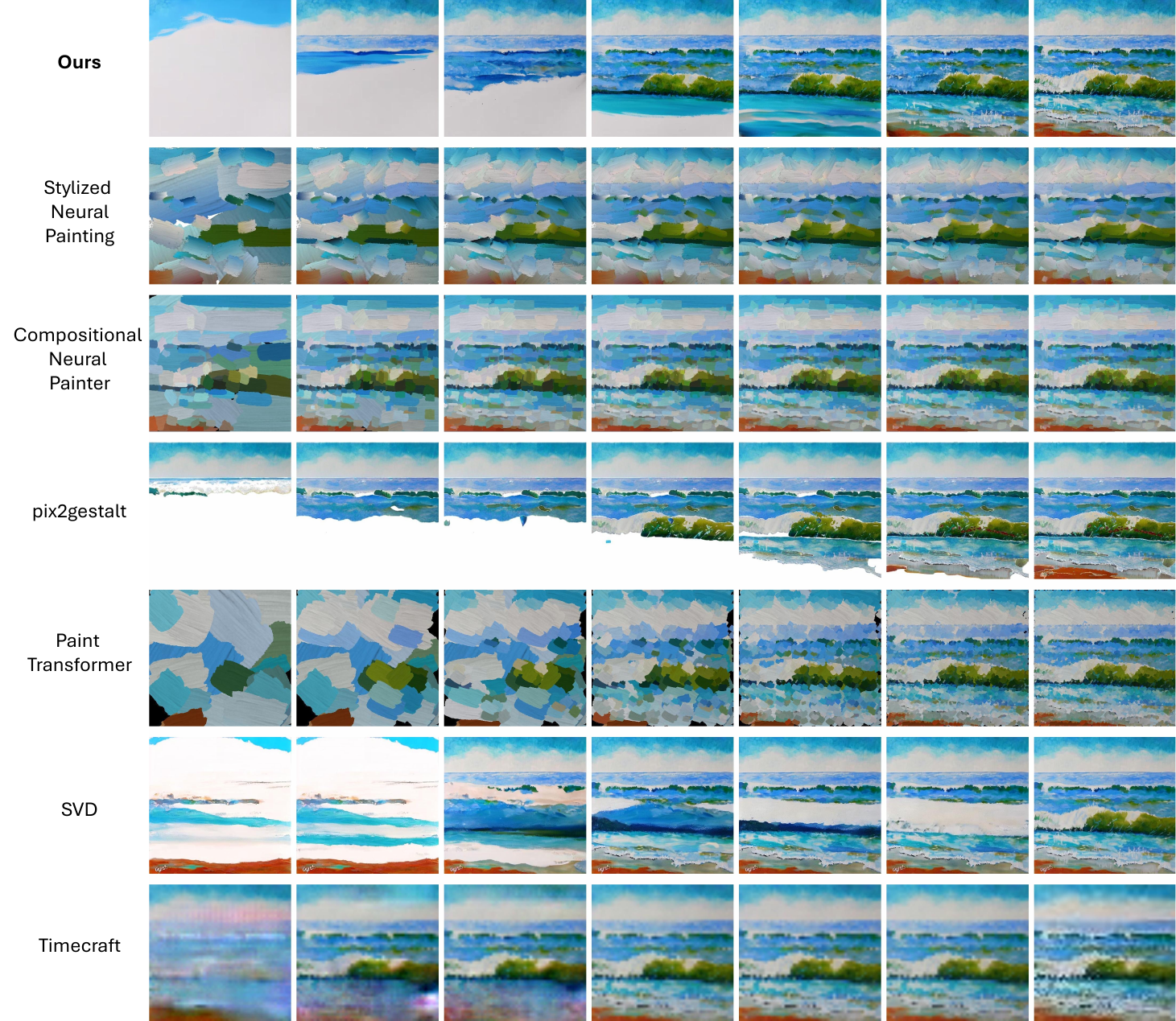}
    \caption{ \textbf{Comparison with baselines on in-the-wild images.}  Our method significantly outperforms these baselines in generating a more human-like painting video with better visual quality. Image courtesy Catherine Kay Greenup.}
    \label{fig:baseline_comparison1}
\end{figure*}

\begin{figure*}[!t]
    \centering
\includegraphics[scale=0.72 ]{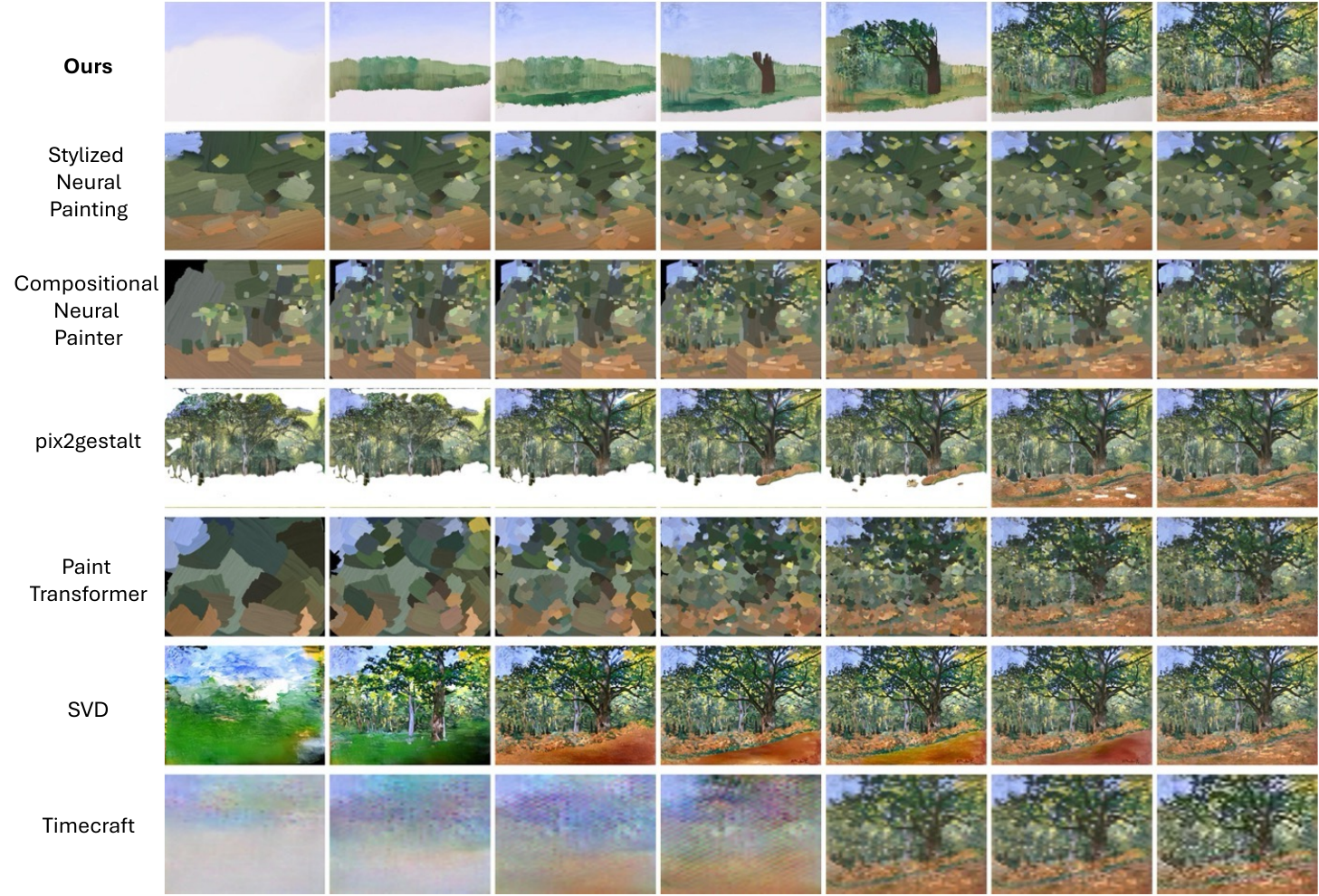}
    \caption{ \textbf{Comparison with baselines on in-the-wild images.}  Our method significantly outperforms these baselines in generating a more human-like painting video with better visual quality. Image courtesy Rawpixel.}
    \label{fig:baseline_comparison2}
\end{figure*}

\begin{figure*}[!t]
    \centering
\includegraphics[scale=0.72 ]{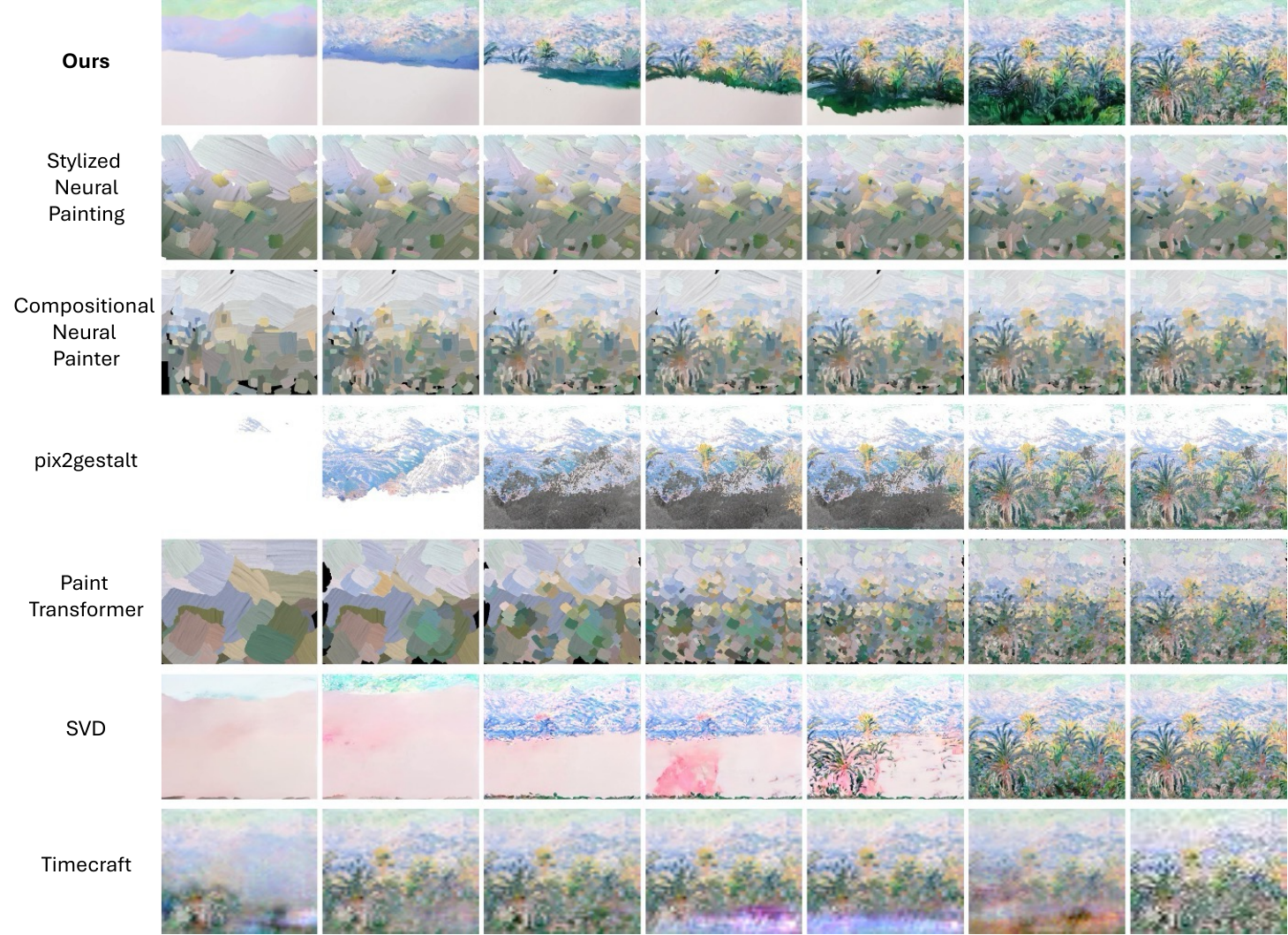}
    \caption{ \textbf{Comparison with baselines on in-the-wild images.}  Our method significantly outperforms these baselines in generating a more human-like painting video with better visual quality. Image courtesy Rawpixel.}
    \label{fig:baseline_comparison3}
\end{figure*}

\begin{figure*}[!t]
    \centering
\includegraphics[scale=0.72 ]{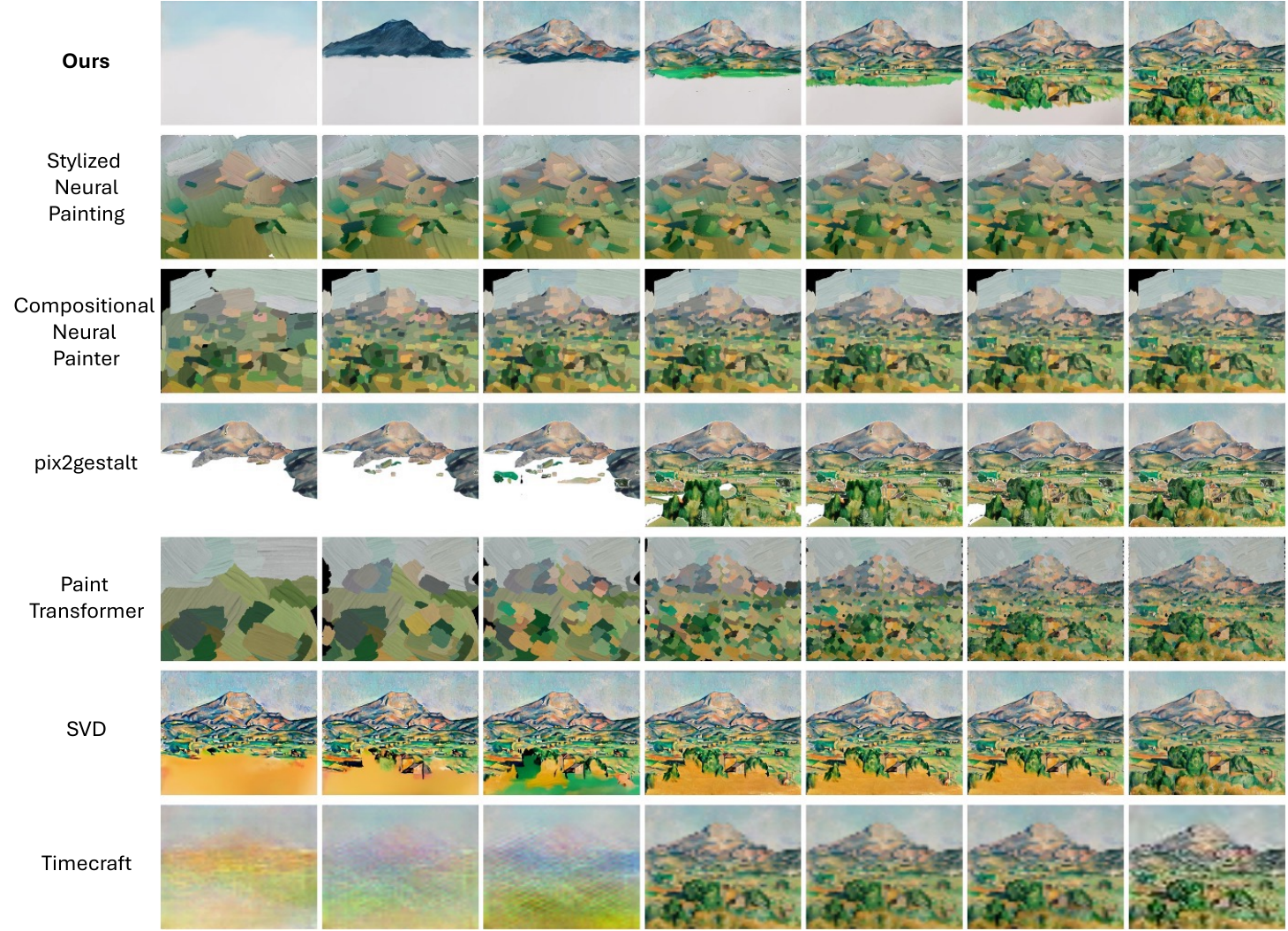}
    \caption{ \textbf{Comparison with baselines on in-the-wild images.}  Our method significantly outperforms these baselines in generating a more human-like painting video with better visual quality. Images courtesy Barnes Foundation.}
    \label{fig:baseline_comparison4}
\end{figure*}

\begin{figure*}[!t]
    \centering
\includegraphics[scale=0.72 ]{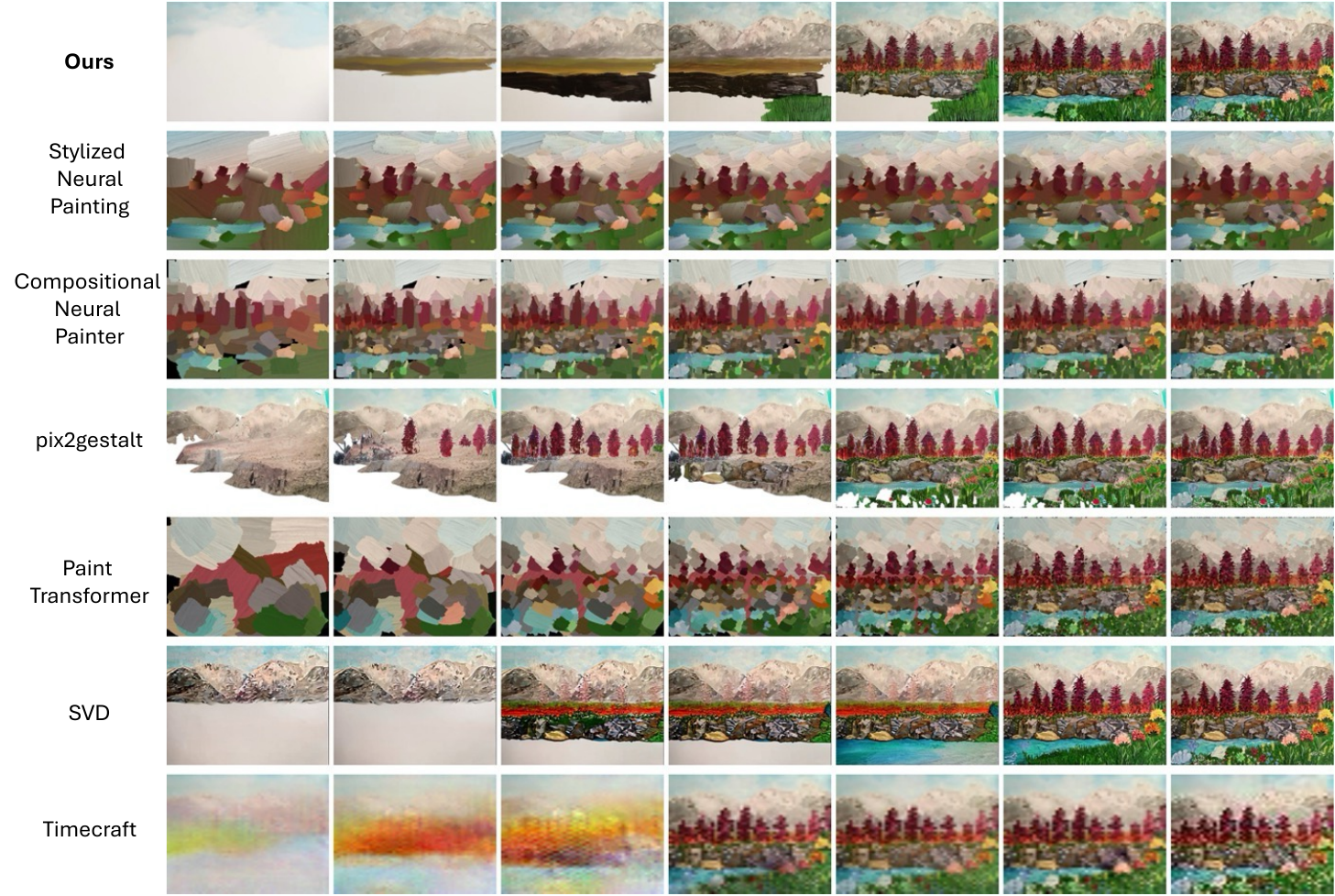}
    \caption{ \textbf{Comparison with baselines on in-the-wild images.}  Our method significantly outperforms these baselines in generating a more human-like painting video with better visual quality. Image courtesy Catherine Kay Greenup.}
    \label{fig:baseline_comparison5}
\end{figure*}

\noindent\textbf{Full Paintings}.
We provide qualitative comparisons with all baselines for full paintings in Fig.~\ref{fig:baseline_comparison1}, Fig.~\ref{fig:baseline_comparison2}, Fig.~\ref{fig:baseline_comparison3},
Fig.~\ref{fig:baseline_comparison4}, and Fig.~\ref{fig:baseline_comparison5}. The three stroke-based rendering baselines -- \textit{Stylized Neural Painting}, \textit{Compositional Neural Painter}, and \textit{Paint Transformer} -- apply brushstrokes in a non-human-like manner, as they are not trained on actual painting videos. 
Furthermore, due to the limitations of parameterized brushstroke constraints, they produce only a ``stylized'' version of the target painting.  
\textcolor{black}{\textit{pix2gestalt}'s results are non-human-like and exhibited visual artifacts. This is due to inaccurate predefined painting order, imperfect segmentation, and unnecessary paints on the canvas to complete unoccluded segments. 
}
\textit{SVD} often gets stuck in the painting process, evident in columns 4 to 6 of Fig.~\ref{fig:baseline_comparison4}, and produces visual artifacts, such as the unreasonable colors in columns 1 and 2 of Fig.~\ref{fig:baseline_comparison2} and columns 1 to 5 of Fig.~\ref{fig:baseline_comparison3}. 
Moreover, despite being trained on a real painting dataset, it still fails to mimic the human painting order reasonably. 
\textit{Timecraft} generates only very low-resolution sequences and introduces noticeable visual artifacts. In contrast, our method significantly outperforms all baselines in mimicking human-like painting sequences, focusing on focal areas, employing layering techniques, and achieving good video quality.

Table.~\ref{tab:crop_quantitative} presents the quantitative comparisons across all baselines and metrics. Our method outperforms all baselines in every evaluated metric.

\begin{figure}[!t]
\centering
  \subcaptionbox{DC of a Single Painting}%
{\includegraphics[width=0.48\linewidth]{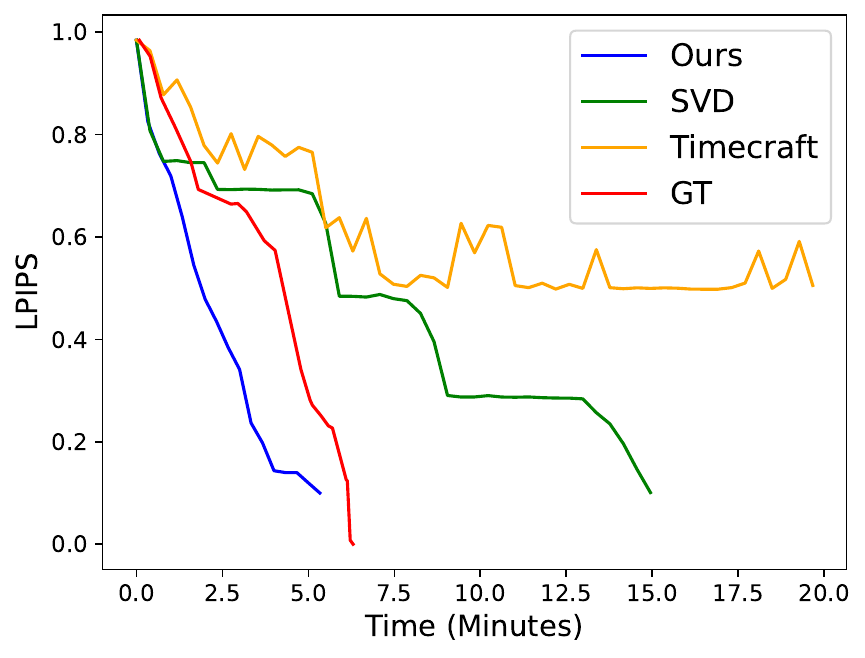}}
  \subcaptionbox{Slope Distribution of DC}[0.48\linewidth]{
    \begin{tikzpicture}
      \node[anchor=south west,inner sep=0] (image) at (0,0) {
        \includegraphics[width=\linewidth]{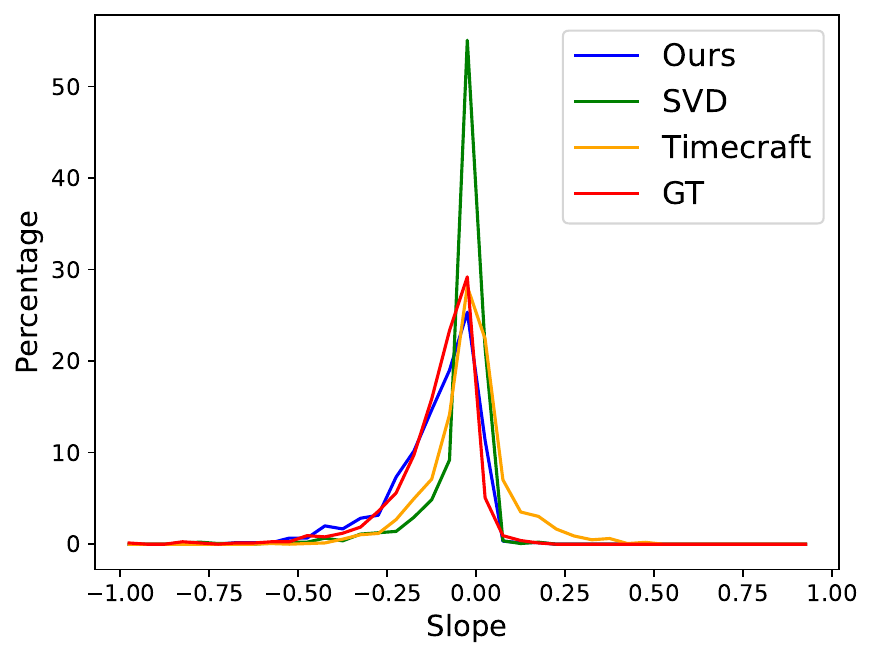}
      };
      \begin{scope}[x={(image.south east)},y={(image.north west)}]
        \draw[-latex, thick, red] (0.7,0.5) -- (0.56,0.48);
      \end{scope}
    \end{tikzpicture}
  }
\caption{\textbf{Quantitative analysis of painting process.}
(a) illustrates the distance curves (DC) of various methods applied to a specific painting; (b) visualizes the distribution of slope of these distance curves across all paintings in the validation set. In (b), we classify the slopes into 20 evenly spaced intervals ranging from -1 to 1, and plot the percentage of the slope value (y-axis) falls into each interval (x-axis).  For instance, the peak of the green curve in (b) shows that over 50 percent of the slopes range from -0.05 to 0. A negative slope value suggests that the update to the canvas bring it closer to the target image (as expected), and vice versa. 
  }
  \label{fig:slope_curve}
\end{figure}

Further, we evaluate how the painting processes generated by various methods progress toward the target painting in terms of speed and direction using the distance curve. 
The distance curve of a sequence depicts its progression towards the target painting, plotting time (x-axis, minutes) against LPIPS distance (y-axis) between target and current images. 
Among baselines, we select \textit{SVD} and \textit{Timecraft} for analysis. 
For these two baselines, we use a time interval of 23.6 seconds, matching the average duration of our training set. 
Fig.\ref{fig:slope_curve} (a) illustrates the distance curve  of various methods applied to a single painting in the validation set. \textit{Timecraft} fluctuates considerably and fails to consistently approach the target image. Besides, it does not converge because its outputs are in a very low resolution.  
\textit{SVD} encounters stalls during the painting process, leading to extended convergence times. 
Our method exhibits the curve that most closely aligns with that of the GT. Note that, neither our method nor \textit{SVD} achieves perfect reconstruction of the target painting due to the utilization of VAE.
For further analysis, Fig.\ref{fig:slope_curve} (b) presents the distribution of the slope of these distance curves across all paintings in the validation set. \textit{Timecraft} exhibits over 20\% of its slopes between 0 to 0.05 (marked with a red arrow), indicating frequent deviations from the target image, such as erasing and adding unrelated colors.
 \textit{SVD}'s slopes, with over 50\% ranging from -0.05 to 0, indicate minimal updates on the canvas. In contrast, our method's distributions are closer to those of the GT, demonstrating that our painting process progresses at a reasonable speed and direction.

\begin{figure}
    \centering
\includegraphics[scale=0.56 ]{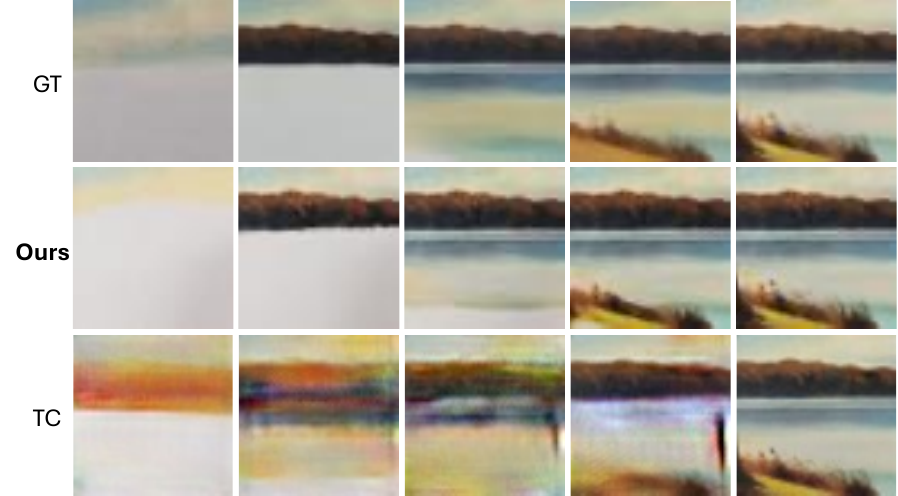}
    \caption{ \textbf{Qualitative comparison on low-resolution painting crops.} We compare with \textit{Timecraft} (TC) on low-resolution painting crops. \textit{Timecraft} produces artifacts and fails to produce a human-like painting process. In contrast, our method delivers a more realistic painting video with better quality. }
    \label{fig:baseline_comparison_crop}
\end{figure}

\noindent\textbf{Cropped Paintings}.
We compare with \textit{Timecraft} by randomly selecting 3 low-resolution crops from every downsampled full painting in the validation set, following the cropping strategy presented in the paper of  \textit{Timecraft}. Fig.~\ref{fig:baseline_comparison_crop} provides a qualitative comparison of cropped paintings with \textit{Timecraft}, where our approach yields a more authentic painting process in terms of painting order, focus regions, and overall video quality. The quantitative results in Table.~\ref{tab:crop_quantitative} further demonstrate that our method outperforms \textit{Timecraft} across all metrics.

\begin{figure*}[!t]
\centering
  \subcaptionbox{Current Image}%
{\includegraphics[width=0.245\linewidth]{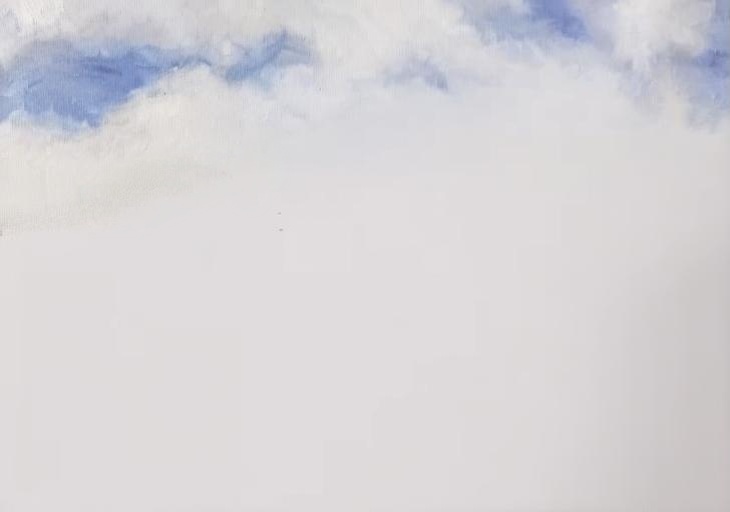}}
  \subcaptionbox{Ours-CE}%
{\includegraphics[width=0.245\linewidth]{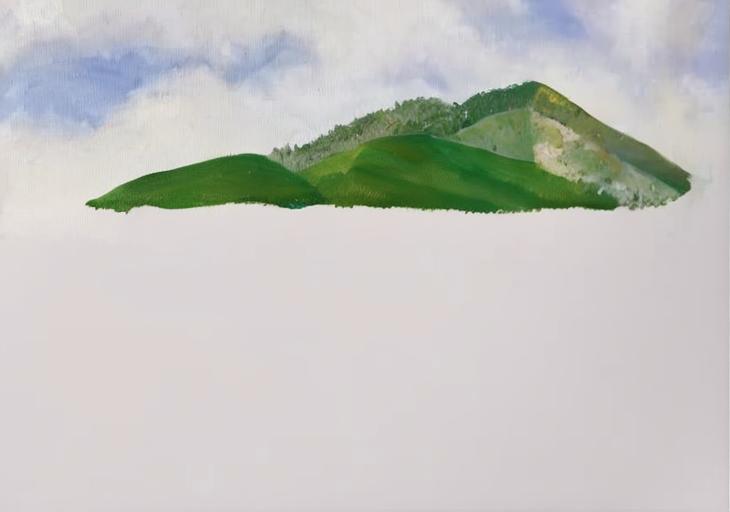}}
  \subcaptionbox{Ours}%
{\includegraphics[width=0.245\linewidth]{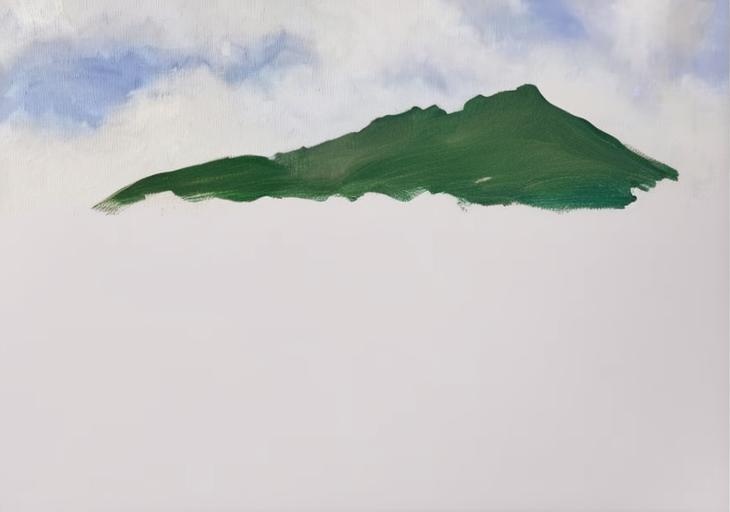}}
\raisebox{-5.5mm}{\rule{1pt}{36.5mm}}%
\hspace{1pt}
  \subcaptionbox{Target Image}%
{\includegraphics[width=0.245\linewidth]{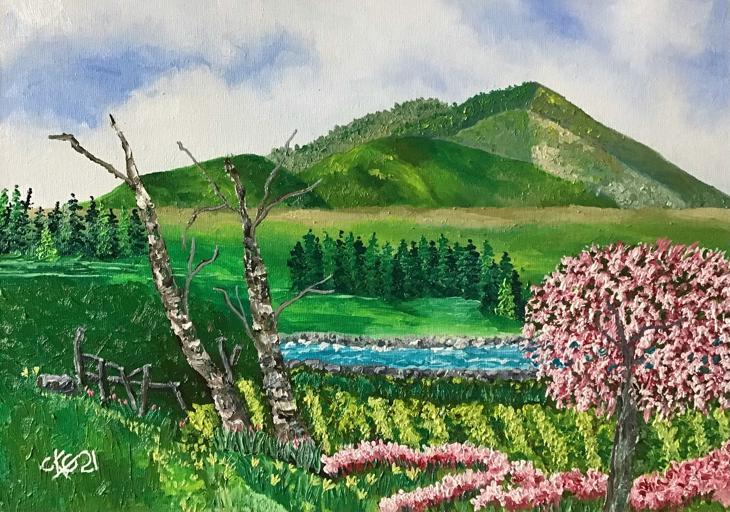}}
\caption{\textcolor{black}{\textbf{Ablation of the predicted CLIP embeddings.} The current and target images are shown in (a) and (d), respectively. (b) and (c) represent the outputs generated by different models at the same timestep, utilizing the same time interval, current image, and target image.
Without incorporating predicted CLIP embeddings (b), the model completes the mountain in full detail.  The process involves frequent switching of color brushes, deviating from the painting style observed in the training set. Our full model (c) paints the base layer of the mountain first and leaves the details for latter stages, which follows the artistic techniques presented by the artists in our training data.  Please see Fig. 6 in the main paper for more keyframes generated by the full model. Image courtesy Catherine Kay Greenup.  }
}
  \label{fig:clip}
\end{figure*}

\subsection{Ablation Study}

\begin{figure*}[!t]
    \centering
\includegraphics[scale=0.73]{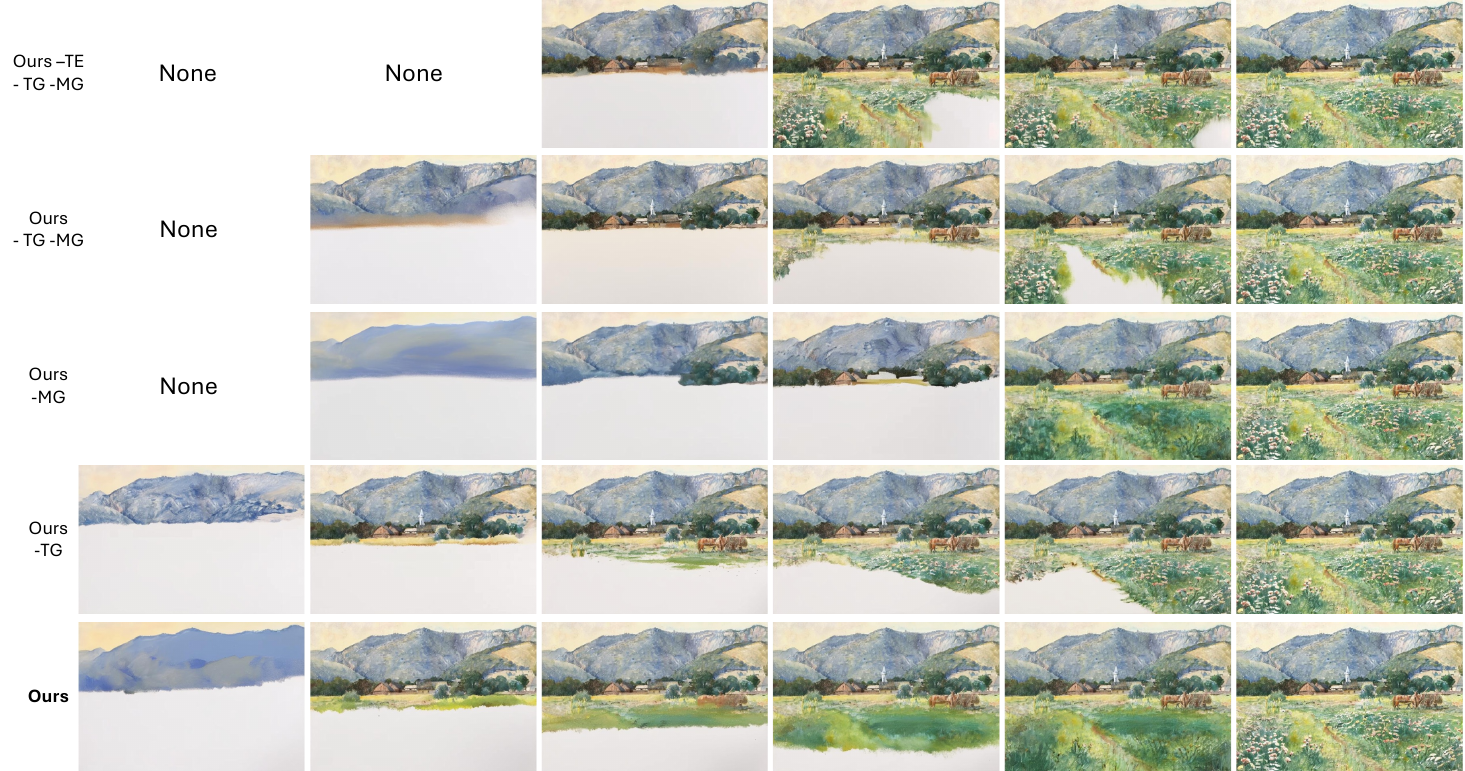}
    \caption{ \textbf{Ablation study of different conditional signals.} The row with ``None'' means there are no enough keyframes for display. For instance, the first row has only four keyframes, leaving the first two columns ``None''.  \textit{Ours-TE-TG-MG} generates excessive content for each update, resulting in only four keyframes throughout the entire painting process.   \textit{Ours-TG-MG} slows generation slightly but still converges quickly. Besides, it also produces unreasonable rendering, as shown in column 2. \textit{Ours-TG}, relying heavily on text instructions, tends to complete entire semantic classes in each update, promoting swift convergence.   \textit{Ours-MG} leverages region masks to moderate the speed of generation. Yet, in the absence of textual guidance, it struggles to adequately differentiate between semantic classes. 
For example, in column 4 to 6, the flower and grass are painted simultaneously, rather than using layering techniques that complete the grass first and add the flower afterward. In contrast, \textit{Ours} achieves a more logical and orderly painting process.   Image courtesy Julius Zorkoczy, Landscape with a Blooming Meadow, Slovenska narodna galeria, SNG, \url{https://www.webumenia.sk/dielo/SVK:SNG.K_1710}.   
}
    \label{fig:ablation}
\end{figure*}

In Fig.~\ref{fig:ablation}, we present the ablation studies for variants omitting different conditional signals. Both one-step variants, \textit{Ours-TE-TG-MG} and \textit{Ours-TG-MG}, yield unsatisfactory results, underscoring the significance of the two-stage design. \textit{Ours-MG} heavily depends on text instructions and tends to complete an entire semantic class with each update, unexpectedly accelerating the generation process excessively. \textit{Ours-TG} struggles to comprehend the semantic contents of the target paintings, consequently painting the grass and flowers simultaneously without employing layering techniques. 
\textcolor{black}{We further present qualitative results of the variants without considering predicted CLIP embedding (\textit{Ours-CE}) in Fig.~\ref{fig:clip}. It completes details of the mountain at an early stage, which fails to mimic the painting style of artists in our training set. }
In contrast, \textit{Ours} delivers the most human-like painting process, characterized by a logical painting order, targeted focal regions, and proper use of layering techniques.

\subsection{Human Study}

As described in the main paper, we normalized the ratings to remove user bias. Specifically, we divided each participant's rating for a specific painting sequence by the sum of their ratings for all 4 painting sequences (of the same target painting). We then averaged these normalized ratings across all paintings and participants for each method.

Our method achieved the highest average normalized rating, surpassing \textit{SVD} by 1.9 times, \textit{Paint Transformer} by 2.1 times, and \textit{Timecraft} by 3.0 times  (reported in the main paper). 
Without normalization, our method received the highest rating at 4.21, compared to \textit{SVD} (2.96), \textit{Paint Transformer} (2.11), and \textit{Timecraft} (1.52).

\subsection{Influence of Random Seeds}

\begin{figure*}[!t]
    \centering
\includegraphics[scale=0.73]{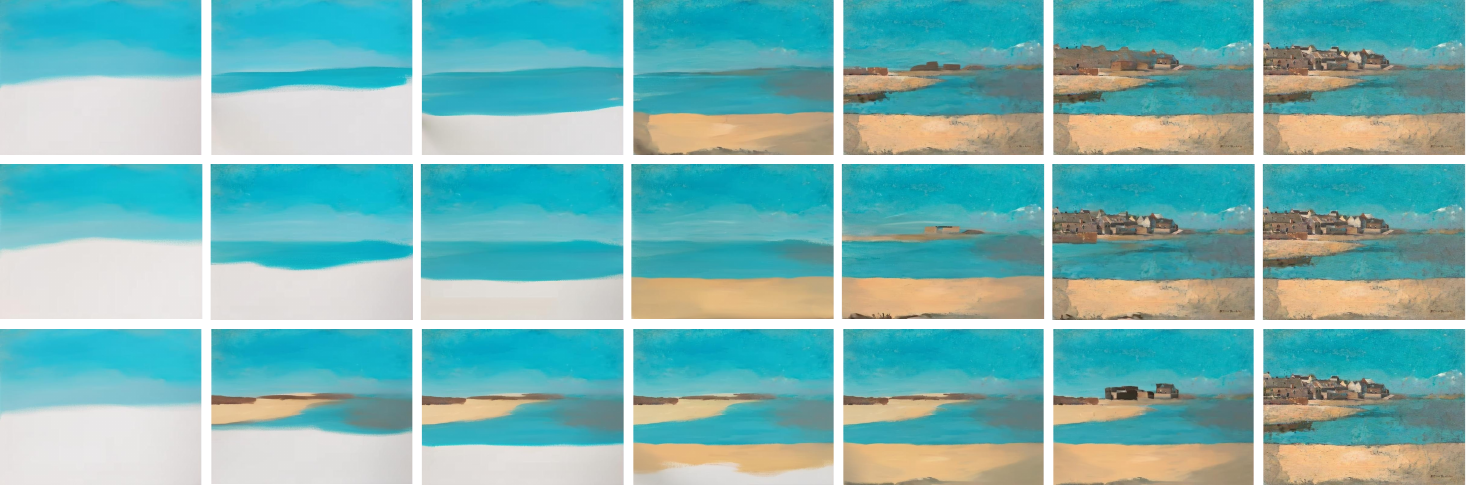}
    \caption{ \textbf{Comparison of using different random seeds for the diffusion model in our method.} Each row displays the results of a specific random seed. Despite using different random seeds, the painting orders are generally similar (i.e., back to front) with slight differences in painting foreground objects. For example, rows 1 and 2 tend to address the far ground and house in the final stages, whereas row 3 completes the far ground immediately after finishing the background nearby. Both styles of painting order are observed in the training set, and our method successfully captures these variations. Image courtesy National Gallery of Art. }
    \label{fig:random_seed}
\end{figure*}

In Fig.~\ref{fig:random_seed}, we illustrate the outcomes of utilizing different random seeds for the diffusion model in our methods. Although different seeds are used, the generated painting processes generally follow a similar order, with minor variations in the sequence of painting foreground objects. These variations are reflective of those observed in the training data, demonstrating that our method can learn the general painting order from these slightly varied sequences and capture the variability. This adaptability allows for the use of different random seeds at inference time to achieve diverse results.

\subsection{Failure Case}

\begin{figure}[!t]
    \centering
\includegraphics[scale=0.27]{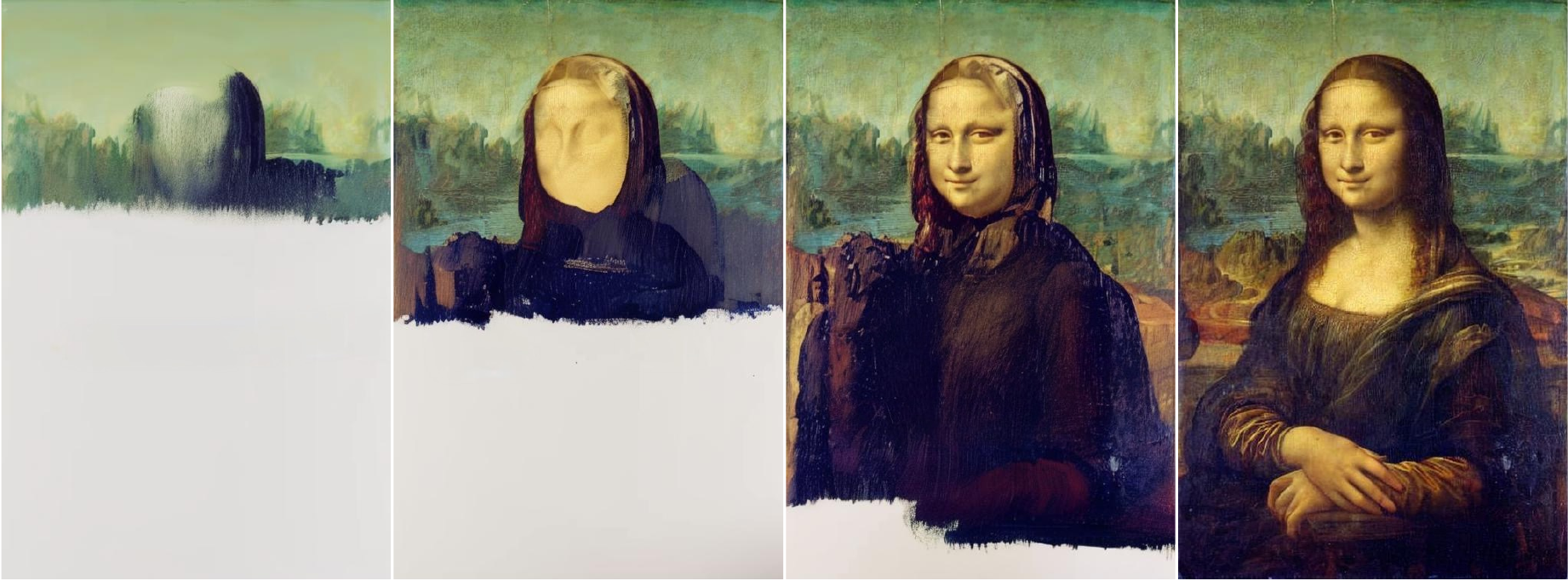}
    \caption{\textbf{More failure cases.} Trained on landscape paintings, our method struggles with portrait paintings. Image courtesy Rawpixel.  }
    \label{fig:failure}
\end{figure}

Fig.~\ref{fig:failure} shows another failure case of our method on the portrait painting, Mona Lisa.

\newpage

\clearpage
\newpage
\bibliographystyle{ACM-Reference-Format}
\bibliography{sample-bibliography}